%% file: main.tex
\documentclass[runningheads]{llncs}

\usepackage[utf8]{inputenc}
\usepackage{graphicx} 
\usepackage{multirow}
\usepackage{multicol} 
\usepackage{booktabs}
\usepackage{amssymb}
\usepackage{pifont}
\usepackage{textcomp}

\usepackage{caption}
\usepackage{subcaption}
\usepackage{algorithm}
\usepackage{algpseudocode}
\usepackage{bm}
\usepackage{amsmath}
\usepackage{mathtools}
\usepackage{commath}
\usepackage{nomencl}

\usepackage{tikz}
\usepackage{pgf}
\usepackage{pgfplots}
\usepackage{standalone}
\usepackage{bbm}
\usepackage{threeparttable}
\usepackage{hyperref}
\usepackage{soul}
\usepackage{xspace}

\usepackage{float}

\graphicspath{{figs/}}

\newcommand\numberthis{\addtocounter{equation}{1}\tag{\theequation}}

\begin{document}
\title{HeMI: Multi-view Embedding in Heterogeneous Graphs}
%

\author{Costas Mavromatis \and George Karypis}
\institute{Department of Computer Science and Engineering, University of Minnesota, \\ Minneapolis, MN 55454, USA \\\email{mavro016@umn.edu}, \email{karypis@umn.edu}\\ }

\maketitle    
\begin{abstract}
Many real-world graphs involve different types of nodes and relations between nodes, being heterogeneous by nature.
The representation learning of heterogeneous graphs (HGs) embeds the rich structure and semantics of such graphs into a low-dimensional space and facilitates various data mining tasks, such as node classification, node clustering, and link prediction. In this paper, we propose a self-supervised method that learns HG representations by relying on knowledge exchange and discovery among different HG structural semantics (meta-paths). Specifically, by maximizing the mutual information of meta-path representations, we promote meta-path information fusion and consensus, and ensure that globally shared semantics are encoded. By extensive experiments on node classification, node clustering, and link prediction tasks, we show that the proposed self-supervision both outperforms and improves competing methods by 1\% and up to 10\% for all tasks.

\end{abstract}
\section{Introduction}

Heterogeneous graphs (HGs) model compositions of different types of entities and relations and naturally emerge in various real-world applications. Such examples include bibliographic networks, social networks, and recommendation systems. Encoding this high-dimensional, non-Euclidean, and heterogeneous information to a low-dimensional Euclidean embedding space facilitates various data related tasks that benefit from operating on Euclidean spaces, such as classification, clustering, and link prediction. Such an encoding could be done, for example, by using traditional machine learning techniques, which have been proven to be powerful when learning embeddings (a.k.a. representations). However, the graph nature (non-Euclidean structure) and heterogeneity of HGs pose great challenges to their representation learning. Moreover, embedding HGs based on self-supervision signals allows the re-usability of the representations for various tasks and fine-tuning on specific tasks with few labeled data. 

When dealing with homogeneous graphs, different self-supervised graph embedding methods have been proposed for learning representations. Early approaches rely on random walk techniques (e.g., DeepWalk~\cite{perozzi2014deepwalk}) and they mainly preserve the input structure, i.e., nodes that are close in the input should have similar representations. The limitation of these methods is that they are not inherently designed to handle additional feature data of nodes, which are present in most real-world graphs. As a result, graph neural networks (GNNs) have been proposed~\cite{kipf2016semi}, which are deep learning methods designed for handling both the graph structure and the node attributes. GNNs mainly estimate node representations through a recursive neighborhood aggregation scheme. Since GNN already capture the structural information, recent graph embedding methods learn the representations to encode different scales of the data (such as clusters of nodes, e.g., GIC~\cite{mavromatis2020graph}), which conveys higher order information. 

A natural approach of HG representation learning is to extend the existing homogeneous-designed methods such as they handle the heterogeneity of the data. A common way is to model different compositions of HG relations as different meta-paths~\cite{sun2011pathsim}, which then, can be treated as homogeneous graphs by disregarding intermediate relations. Representations of each meta-path can be encoded and optimized by any self-supervised method for homogeneous graphs (e.g., random walk on each meta-path~\cite{dong2017metapath2vec}, or a GNN encoder with GIC self-supervision). Oftentimes, in order to capture the full semantics of the data, it may be necessary to combine and fuse information across various meta-paths. This is mostly achieved with heterogeneous graph neural networks (HGNNs), which aggregate the different meta-path representations to a compact representation (e.g., HAN~\cite{wang2019han}). In this case, the supervision is directly optimizing the compact representation, which is used for downstream tasks.


\textbf{Contribution}. In this paper, we advocate that relying solely on homogeneous-designed self-supervision signals is not sufficient for learning good representations, since they do not help different meta-paths to learn from each other. We propose a heterogeneous-designed method, HeMI, that allows information exchange and knowledge discovery across meta-paths. Specifically, per-meta-path node representations are aggregated together to a fused representation and these representations are jointly optimized so that their mutual information is maximized. As a result, the fused representations encode information that is  mutually helpful to all meta-paths and integrate globally shared semantics to each meta-path representation.

We evaluate  HeMI on four standard datasets using node classification, node clustering and link prediction as the downstream tasks. Experiments show that HeMI outperforms both unsupervised and semi-supervised competing approaches in all tasks. HeMi's improvement over the \emph{best} competing method is over 1\%, 2\%, and 3\% in the majority of the datasets for node classification, node clustering and link prediction, respectively. HeMI is also powerful as an augmented loss objective providing better performance by 2\% and 10\% for node classification and link prediction, respectively.

\section{Preliminary}

\begin{figure}[t]
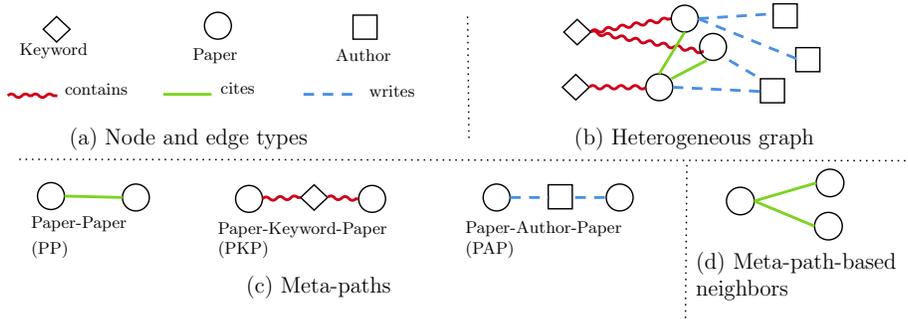

    \centering
    \includestandalone[width=\textwidth]{figs/example}
     \caption{An illustrative example of a heterogeneous graph (citation network) with three types of nodes and three types of edges. }
    \label{fig:example}
\end{figure}

\textbf{Heterogeneous graph}~\cite{sun2013mining}. A heterogeneous graph (HG), denoted as $\mathcal{G} = (\mathcal{V}, \mathcal{E})$, consists of an object set $\mathcal{V}$ and a link set $\mathcal{E}$. It is used to represent a network with
multiple types of nodes $\mathcal{A}$ and links $\mathcal{R}$. A heterogeneous graph is associated with a node type
mapping function  $\chi_\mathcal{V}: \mathcal{V} \xrightarrow{} \mathcal{A}$ and a link type mapping function
$\chi_\mathcal{E}: \mathcal{E} \xrightarrow{} \mathcal{R}$. $\mathcal{A}$ and $\mathcal{R}$ denote the sets of predefined node types and
link types, where $|\mathcal{A}| + |\mathcal{R}| > 2$. The type of a link automatically defines the types of nodes on its two ends.
Fig.~\ref{fig:example}(a) shows an example of a HG where $\mathcal{V}$ consists of \{paper, keyword, author\} nodes and $\mathcal{E}$ consists of \{cites, contains, writes\} links. Here, $\chi_\mathcal{V}$ corresponds to the shape used for illustration of each node type (circle, rectangular, and  diamond) and $\chi_\mathcal{E}$ corresponds to the color used for illustration of each link type (green, blue, and red).

\textbf{Meta-path}~\cite{sun2011pathsim}. A meta-path $P$ is defined as a path in the form $A_1 \xrightarrow{R_1} A_2 \xrightarrow{R_2} \cdots \xrightarrow{R_l} A_l$ (abbreviated as $A_1A_2 \cdots A_l$), which describes a composite relation $R = R_1 \circ R_2 \cdots \circ R_l $ between node types $A_1$ and $A_l$, where $\circ$ denotes the composition operator on relations.
In Fig.~\ref{fig:example}(c), example meta-paths include  
\begin{itemize}
\item Paper $\xrightarrow[]{\text{cites}}$ Paper (PP)
\item Paper $\xrightarrow[]{\text{contains}}$ Keyword $\xrightarrow[]{\text{contained}}$ Paper (PKP)
\item Paper $\xrightarrow[]{\text{written}}$ Author $\xrightarrow[]{\text{writes}}$ Paper (PAP).
\end{itemize}
Different meta-paths reveal different semantics of the data. Note that we could obtain more meta-paths, such as PAPK, PKPPA, etc. 

\textbf{Meta-path based neighbor}. 
Given a meta-path $P$ whose starting vertices are of type $A_1$ and ending vertices are of type $A_l$, the meta-path based neighbors $\mathcal{N}^P_v$ of $v \in A_1$ are defined as the set of $A_l$ nodes that connect with $v$ via meta-path $P$ (following the right type of relations). 
Note that the node’s neighbors include itself. 
Fig.~\ref{fig:example}(d) shows the meta-path based neighbors of the upper paper node via meta-path PP. 

\textbf{Meta-path based graph}. Given a meta-path $P$ whose starting vertices are of type $A_1$ and ending vertices are of type $A_l$, the meta-path based graph $\mathcal{G}^P$ is a graph constructed from all node pairs $v \in A_1$ and $u \in A_l$ that connect via meta-path $P$. Meta-path based graphs can exploit different aspects of structure information in heterogeneous graph.

\textbf{Heterogeneous graph representations}. Given a
heterogeneous graph $\mathcal{G} = (\mathcal{V}, \mathcal{E})$, with node attribute matrix $\mathbf{X} \in \mathbb{R}^{|\mathcal{V}| \times d_{\text{in}}}$, heterogeneous graph embedding is the task of learning the $d$-dimensional node representations $\mathbf{z}_v \in \mathbb{R}^d$ for all $v \in \mathcal{V}$. It is desired that the representations are able to capture rich structural and semantic information involved in $\mathcal{G}$. The learned (low-dimensional) representations can be applied to downstream graph-related tasks such as node classification, node clustering, and link prediction. We focus on learning the representations of specific types of nodes in this paper, i.e., $\mathbf{Z} \in \mathbb{R}^{|\mathcal{V}_t| \times d}$, where $\mathcal{V}_t$ is the set of nodes of the target types. 

Here, we denote vectors by bold lower-case letters and matrices by bold upper-case letters. We also use the terms \emph{representation} and \emph{embedding} interchangeably.

\section{Related Work}\label{sec:rel}

\textbf{Homogeneous graph representation learning}.
Early graph representation learning approaches, such as DeepWalk (DW)~\cite{perozzi2014deepwalk},  optimize the embeddings so that nodes that tend to co-occur on short random walks over the graph have similar embeddings. Following the paradigm of DeepWalk, node2vec (N2V)~\cite{grover2016node2vec} and LINE~\cite{tang2015line} are more generalized versions of DeepWalk with biased random walks and additional proximity measures, respectively. Since these methods are not inherently designed to handle node attributes, GNNs have been proposed to encode both the structural and semantic information of the graph and its attributes. Widely used GNNs include GCN~\cite{kipf2016semi}, which at each layer learns to convolve features (or embeddings) of 1-hop neighbors, and GAT~\cite{velickovic2018graph}, which substitutes the statically normalized convolution operation of GCN with an attention mechanism. Leveraging GNNs to handle the structural properties of the graph, recent self-supervised methods do not rely on structure-based proximities, but optimize the representations based on higher order similarities. DGI~\cite{velickovic2018deep} encodes information that is shared among all nodes of the graph, while GIC~\cite{mavromatis2020graph} extends DGI to encode additional shared information within clusters of nodes. In a similar manner, MVGRL~\cite{hassani2020contrastive} uses data augmentation techniques to obtain different views of the graph and to encode the information that is shared among these views.

\noindent
\textbf{Heterogeneous graph representation learning}. Inspired by homogeneous graph representation learning, most heterogeneous based approaches handle and encode the heterogeneity of the graph by operating on meta-paths, which can also be treated as homogeneous graphs. For example, metapath2vec (M2V)~\cite{dong2017metapath2vec} and ESim~\cite{shang2016meta} are random walk methods applied on a single meta-path and multiple meta-paths, respectively. Unlike ESim, that cannot learn the importance of each meta-path, HERec~\cite{shi2018herec} utilizes a fusion algorithm to better combine information from different meta-paths. The fusion strategy inspired the development of HGNNs to better handle the heterogeneity of the graph compared to conventional GNNs. For example, HAN~\cite{wang2019han} uses a GAT-like encoder on each meta-path and learns to aggregate the information from these meta-paths with an additional attention mechanism. This strategy can be applied to any homogeneous self-supervised method to handle the heterogeneity of the graph; for example, we later extend GIC to heterogeneous GIC (HGIC). HDGI~\cite{ren2019hdgi} acts in similar manner to obtain compact representations, which are optimized with the DGI objective. DMGI~\cite{park2020dmgi} enhances the DGI objective with a consensus regularization so that meta-path representations are similar to each other.   Different from most methods that treat each meta-path as a homogeneous graph, MAGNN~\cite{fu2020magnn} extends HAN by considering both the meta-path based neighborhood and the nodes along the meta-path, so that all the meta-path information is preserved for learning. Finally, since most heterogeneous graphs can be treated as knowledge graphs and vice versa,  many learning methods on knowledge graphs, e.g., RGCN~\cite{schlichtkrull2018rgcn} and RotatE~\cite{sun2019rotate}, can often be applied to heterogeneous graph learning. Applications of HGNNs range from citations networks~\cite{wang2019han} and recommendation systems~\cite{shi2018herec} to biological networks~\cite{shui2020hmgnn} and drug interactions~\cite{ioannidis2020panrep}.

\section{Methodology}

\subsection{Hypothesis and motivation}

We assume that the HG $\mathcal{G}$ is given as distinct meta-path graphs $\mathcal{G}^{P_1}, \dots, \mathcal{G}^{P_M}$, which can be regarded as different views (multi-views) -- each describing a different perspective. 
We also assume that all meta-paths contain useful information that needs to be encoded (the meta-paths could be provided by experts or data engineers) and our methodology is postulated on the following: 

First, specific knowledge of one view can be exploited by other views in order to better describe the underlying data and unveil hidden semantics. Thus, \emph{meta-path fusion}, i.e., information exchange, among views seems necessary in order to obtain good representations. 
Second, good representations are the ones that simultaneously benefit all meta-paths. Thus, \emph{meta-path consensus} among the views is needed in order to decide which information is globally helpful and to prevent over-emphasizing or neglecting information associated with specific meta-paths.


For the example in Fig.~\ref{fig:example}, suppose we wish to make recommendations based on a particular paper. Then, a natural approach is to regard simultaneously which papers it cites, which keywords it contains, and by whom authors it is written (meta-path fusion). Since not all citations, keywords or authors are equally important (e.g., out-dated prior work, keywords co-used for multiple domains, co-authors of different disciplines), a good representation of the paper is the one that preserves the information that best describes its overall contribution (meta-path consensus). By computing such representations, good recommendations include papers with similar scientific contributions.

\begin{figure}[t]
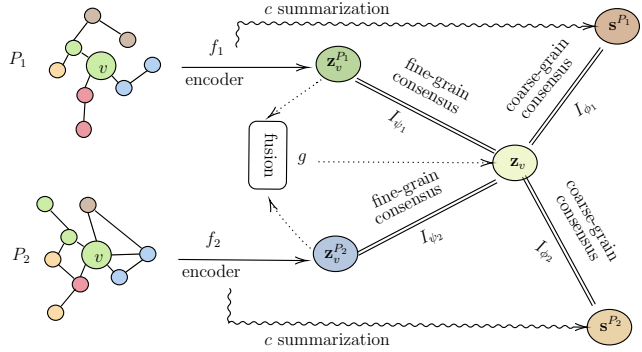

    \centering
    \includestandalone[width=0.7\textwidth]{figs/framework}
    \caption{HeMI framework. The representations for node $v$ for meta-path $P_1$, $\mathbf{z}^{P_1}_v$, and for meta-path $P_2$, $\mathbf{z}^{P_2}_v$, are computed with encoders $f_1$ and $f_2$, respectively. The representations are aggregated with a function $g$ to the fused representation $\mathbf{z}_v$. In addition, nodes' representation in $P_1$ and $P_2$ are summarized to $\mathbf{s}^{P_1}$ and $\mathbf{s}^{P_2}$, respectively. All $\mathbf{z}^{P_1}_v, \mathbf{s}^{P_1}, \mathbf{z}^{P_2}_v$, and $\mathbf{s}^{P_2}$, along with $\mathbf{z}_v$, are optimized by maximizing their consensus, i.e., mutual information, with $\mathbf{z}_v$. Different MI estimators, $\mathcal{I}_{\psi_1}, \mathcal{I}_{\phi_1}, \mathcal{I}_{\psi_2},$ and $\mathcal{I}_{\phi_2}$, are employed for each consensus case.} 
    \label{fig:framework}
\end{figure}



\subsection{A mutual information maximization solution}
In this subsection, we present our proposed framework (a graphical illustration is given in Fig.~\ref{fig:framework}). 

For each meta-path graph $\mathcal{G}^{P_j}$ we capture its structure and attributes by employing a  GNN/HGNN encoder $f_j$ to obtain representations for each node $v$; i.e., $\mathbf{z}^{P_j}_v = f_j(\mathcal{G}^{P_j}_v)$. 
In order to achieve meta-path information fusion, we use an aggregation function $g: \mathbb{R}^{|\mathcal{V}_t| \times d} \times \dots \times  \mathbb{R}^{|\mathcal{V}_t| \times d} \xrightarrow{} \mathbb{R}^{|\mathcal{V}_t| \times d}$  which combines and compresses the representations $\mathbf{z}^{P_j}_v$ together to  $\mathbf{z}_v$ such as $\mathbf{z}_v = g(\mathbf{z}^{P_1}_v, \dots, \mathbf{z}^{P_M}_v)$ (for simplicity, we assume the output dimensionality is the same). 

The representations are optimized by focusing on achieving meta-path consensus. However, instead of achieving consensus between all meta-path combinations (which exponentially grow to the number of the meta-paths), we use $\mathbf{z}_v$ to globally guide the consensus between the meta-paths. In order to measure the consensus, we use the mutual information (MI) $\mathcal{I}$ between $\mathbf{z}_v$ and each one of the meta-path specific representations; i.e., $\mathcal{I}(\mathbf{z}^{P_j}_v; \mathbf{z}_v)$. We maximize the consensus between all $\mathbf{z}^{P_j}_v$ by maximizing the MI of each $\mathbf{z}^{P_j}_v$ with $\mathbf{z}_v$ to achieve global optimization, such as 
\begin{equation}
   \max \; \sum_j \mathcal{I}(\mathbf{z}^{P_j}_v; \mathbf{z}_v).
\end{equation}

The MI measure has several advantages. If the encoders $f_j$ are injective~\cite{xu2018powerful}, we have $\mathcal{I}(\mathcal{G}^{P_j}_v; \mathbf{z}_v) = \mathcal{I}(\mathbf{z}^{P_j}_v; \mathbf{z}_v)$, and thus can achieve MI maximization with the actual input; here, $\mathcal{G}^{P_j}_v$ is a subgraph centered around node $v$. Moreover, if a meta-path contains a lot of noise, this does not increase the MI and thus, is not preferred to be encoded; in contrast to other measures, such as Euclidean distances, which are affected by noisy instances. 

Since mutual information is in general intractable, it is often estimated by a parametrized function $\mathcal{I}_\psi$, such as a deep neural network. The estimated MI is a lower bound of the true MI, which is maximized during optimization in the course of maximizing the true MI. The MI estimation becomes 
\begin{equation}
    \sum_j\mathcal{I}(\mathbf{z}^{P_j}_v,\mathbf{z}_v) \geq \sum_j\mathcal{I}_{\psi_j}(\mathbf{z}^{P_j}_v;\mathbf{z}_v).
    \label{eq:lb1}
\end{equation}

In self-supervised settings for graphs, it is common to maximize the MI between representations and coarse-grain parts of the graph, such as the full graph representation~\cite{velickovic2018deep} or representations of clusters of nodes~\cite{mavromatis2020graph}. This facilitates to propagate higher order properties to the representations. We follow a similar strategy and compute the summary of each meta-path graph as $\mathbf{s}^{P_j} = c(\mathbf{Z}^{P_j})$, where $c: \mathbb{R}^{|\mathcal{V}_t| \times d} \xrightarrow{ } \mathbb{R}^{1 \times d}$. Due to the data processing inequality of MI and then by using MI estimators, we have  
\begin{equation}
    \mathcal{I}(\mathbf{z}^{P_j}_v; \mathbf{z}_v) \geq \mathcal{I}(\mathbf{s}^{P_j}; \mathbf{z}_v) \geq \mathcal{I}_{\phi_j}(\mathbf{s}^{P_j};\mathbf{z}_v).
    \label{eq:lb2}
\end{equation}

Eq.~\eqref{eq:lb1} and Eq.~\eqref{eq:lb2} are two lower bounds of the desired MI $\sum_j \mathcal{I}(\mathbf{z}^{P_j}_v; \mathbf{z}_v)$, which both can be maximized.
By multiplying each bound with $\lambda$ and $(1-\lambda)$, respectively, and adding them together, we obtain the final objective

\begin{equation}
   \max \sum_j \lambda \mathcal{I}_{\psi_j}(\mathbf{z}_v^{P_j},\mathbf{z}_v) + (1-\lambda) \mathcal{I}_{\phi_j}(\mathbf{s}^{P_j},\mathbf{z}_v),
   \label{eq:obj}
\end{equation} 
where $\lambda \in \mathbb{R}$ controls the relative importance of each objective term.

Our proposed method is designed to compute embeddings that optimize Eq.~\eqref{eq:obj}.
Since our method relies on \underline{He}terogeneous graph \underline{MI} maximization, we name our method \underline{HeMI}. 

\subsection{Implementation}

\textbf{MI estimators}. In order to estimate the MI we use the Jensen-Shannon MI estimator. The estimator's objective maximizes the expected $\log$-ratio of the samples from the joint distribution (positive examples) and the product of marginal distributions (negative examples). It acts as a binary cross-entropy (BCE) loss between positive and negative examples with an aid of a function $T_\psi: \mathbb{R}^d \times \mathbb{R}^d \xrightarrow{} \mathbb{R}$ with learnable parameters $\psi$,  which assigns higher scores to the positive than the negative examples. The objective for each $v \in \mathcal{V}_t$ is
\begin{equation}
\min \; \lambda \mathcal{L}_f + (1-\lambda) \mathcal{L}_c, 
\label{eq:loss}
\end{equation}
where
\begin{align*}
    \mathcal{L}_f = & \frac{1}{M} \sum_j \Big( \log T_{\psi_j}(\mathbf{z}^{P_j}_v, \mathbf{z}_v) + \log \big(1 - T_{\psi_j}(\mathbf{z}^{P_j}_v, \tilde{\mathbf{z}}_v) \big) \Big), \numberthis
    \label{eq:fine}
    \\
      \mathcal{L}_c = & \frac{1}{M} \sum_j \Big( \log T_{\phi_j}(\mathbf{s}^{P_j}, \mathbf{z}_v) + \log \big(1 - T_{\phi_j}(\mathbf{s}^{P_j}, \tilde{\mathbf{z}}_v) \big) \Big). \numberthis
    \label{eq:coarse}
\end{align*}

Positive examples are pairings of $(\mathbf{z}^{P_j}_v, \mathbf{z}_v)$ and $(\mathbf{s}^{P_j}, \mathbf{z}_v)$, and negatives are pairings of $(\mathbf{z}^{P_j}_v, \tilde{\mathbf{z}}_v)$ and $(\mathbf{s}^{P_j}, \tilde{\mathbf{z}}_v)$. We obtain $\tilde{\mathbf{z}}_v$ as the output of our model with a corrupted input, which in our case is the original graph with row-shuffled features (for node $v$ we use the attributes of a randomly selected node). As functions $T_{\psi_j}, T_{\phi_j}$, we use a bilinear scoring function, followed by a logistic sigmoid nonlinearity $\sigma(\cdot)$, which converts scores into probabilities, as $T_{\psi_j}(\mathbf{z}^{P_j}_v, \mathbf{z}_v) = \sigma(\mathbf{z}^{P_j}_v \mathbf{W}_{\psi_j} \mathbf{z}_v)$ and $T_{\phi_j}(\mathbf{z}^{P_j}_v, \mathbf{z}_v) = \sigma(\mathbf{s}^{P_j} \mathbf{W}_{\phi_j} \mathbf{z}_v)$, where $\mathbf{W}_{\psi_j}, \mathbf{W}_{\phi_j} \in \mathbb{R}^{d \times d}$ are learnable weights.  Finally, the summary of each meta-path is computed as  $\mathbf{s}^{P_j} = \sigma \big( \frac{1}{|\mathcal{V}_t|} \sum_v \mathbf{z}^{P_j}_v \big)$.

\noindent
\textbf{Information fusion}. Recall that in order to fuse information among meta-paths, we use an aggregation function $g$ as $\mathbf{z}_v= g(\mathbf{z}^{P_1}_v, \dots,\mathbf{z}^{P_M}_v)$. We employ an attention based function $g$, which is common in the literature~\cite{wang2019han,fu2020magnn} to learn the importance of each meta-path. Specifically, we first compute 
\begin{equation}
    \mathbf{e}^{P_j} = \frac{1}{|\mathcal{V}_t|} \sum_v \mathbf{q}^T ( \mathbf{W}_{\text{sem}} \mathbf{z}^{P_M}_v + \mathbf{b} ),
\end{equation}
where $\mathbf{W}_{\text{sem}} \in \mathbb{R}^{d_m \times d}$, $\mathbf{b} \in \mathbb{R}^{d_m}$ are learnable parameters and $\mathbf{q} \in \mathbb{R}^{d_m}$ is a parametrized attention vector. The importance of each meta-path $\beta^{P_j}$ is obtained by normalizing $\mathbf{e}^{P_j}$ with a softmax function and the representation $\mathbf{z}_v$ is the weighted average of each meta-path graph representation as
\begin{equation}
    \mathbf{z}_v = \sum_j \beta^{P_j} \mathbf{z}^{P_j}_v, \text{ with } \beta^{P_j} = \frac{\exp(\mathbf{e}^{P_j})}{\sum_j \exp(\mathbf{e}^{P_j})}.
\end{equation}

\noindent
\textbf{Encoders}. In order to compute $\mathbf{z}^{P_j}_v$ we can employ both GNN and HGNN encoders. When using GNN encoders, each meta-path is reduced to an adjacency matrix $\mathbf{A}^{P_j} \in \mathbb{R}^{|\mathcal{V}_t| \times |\mathcal{V}_t|}$ with $\mathbf{A}^{P_j}_{vu} = 1$ if nodes $v$ and $u$ connect through meta-path $P_j$, and $\mathbf{A}^{P_j}_{vu} = 0$ otherwise. For example if we use a GCN~\cite{kipf2016semi} encoder, $\mathbf{Z}^{p_j}$ is computed as 
\begin{equation}
    \mathbf{Z}^{p_j} = \sigma( {\mathbf{D}^{P_j}}^{-0.5} \bar{\mathbf{A}}^{P_j}{\mathbf{D}^{P_j}}^{-0.5} \mathbf{X}_t \mathbf{W}^{P_j}_t ),
\end{equation}
where $\bar{\mathbf{A}}^{P_j} = \mathbf{A}^{P_j} + \mathbf{I}_t$ with $\mathbf{I}_t$ the identity matrix, $\mathbf{D}^{P_j}$ is the diagonal node degree matrix of $\bar{\mathbf{A}}^{P_j}$,  $\mathbf{W}^{P_j}_t \in \mathbb{R}^{d_{\mathcal{V}_t} \times d}$ is a learnable matrix, and $\sigma$ is a nonlinear activatios such as PReLU.

HGNN encoders, such as MAGNN~\cite{fu2020magnn}, offer the advantage that they also encode intermediate nodes (of a different type) between two nodes of the same type for each meta-path (intra-meta-path information). Node attributes of different node types may have different dimensions, so first, they are projected to a space with same dimensions as $ \mathbf{X}'_t =\mathbf{X}_t  \mathbf{W}_t $, where $\mathbf{W}_t \in \mathbb{R}^{d_{t} \times d'}$. 
Let $P(v,u)$ be a meta-path instance connecting the target node $v$ and the meta-path-based neighbor $u \in \mathcal{N}^P_v$, the intermediate nodes of $P(v,u)$ are defined as $\{m^{P(v,u)}\} = P(v,u) \text{\textbackslash} \{u,v\}$. An intra-meta-path encoder $g_{P(v,u)}$ transforms all the node features along a meta-path into a single vector, for example at the first layer, as
\begin{equation}
    \mathbf{h}_{P(v,u)} = g_{P(v,u)}\Big( \mathbf{x}'_v, \mathbf{x}'_u, \big\{\mathbf{x}'_k, \forall k \in \{m^{P(v,u)}\}\big\}\Big),
    \label{eq:aggr}
\end{equation}
where $\mathbf{h}_{P(v,u)} \in \mathbb{R}^{d'}$. Various useful $g_{P(v,u)}$ meta-path encoders can be found in~\cite{fu2020magnn}, and the final representation can be obtained by any aggregator (mean, attention, sum, etc.); e.g., the mean aggregator computes $\mathbf{h}^{P}_{v} = \frac{1}{\mathcal{N}^P_v} \sum_{u \in \mathcal{N}^P_v} \mathbf{h}_{P(v,u)}$, followed usually by a nonlinear activation, such as PReLU or ELU.

\section{Experimental Methodology}
In this section, we describe the experimental configuration and evaluation of our and competing methods.
We found out that competing methods treat the same dataset in various different ways to report their performance. In order to show that our method is efficient under different configurations, we employ two different evaluation protocols (EP) as proposed in~\cite{ren2019hdgi} (EP1) and in~\cite{fu2020magnn} (EP2).

\subsection{Datasets}

Four widely used heterogeneous graph datasets from different domains are adapted to evaluate the performance of HeMI. Specifically, we use the ACM, DBLP, IMDB, and Last.fm datasets\footnote{\url{https://dl.acm.org/}, \url{https://dblp.uni-trier.de/}, \url{https://www.imdb.com/}, \url{https://www.last.fm/}}, of which statistics are summarized in Table~\ref{tb:data} and their detailed descriptions in the Supplementary Material. DBLP and IMDB have different configuration for EP1 and EP2. For IMDB, we extract additional labels from the raw data in Section~\ref{sec:hess}, as explained in the Supplementary Material. For nodes without attributes, we assign them one-hot identity vectors.
\addtolength{\tabcolsep}{2pt}   
\begin{table}[t]

\centering
\caption{Datasets.}
\label{tb:data}
\resizebox{\textwidth}{!}{ 
\begin{tabular}{cllccc}
\toprule
Dataset & Node-type: \#Nodes & Edge-type: \#Edges & \#Features & Labels & Meta-paths\\
\midrule
\multirow{3}{*}{ACM} & Paper (P): 3,025 & Paper-Author: 9,744 & 1,870 & P: \textit{Database},  & PAP\\
 & Author (A):  5,835&  Paper-Subject: 3,025 &  & \textit{Wireless  Communication}, & PSP \\
 & Subject (S): 56 & & & \textit{Data  Mining} & \\
 \midrule
 \multirow{5}{*}{DBLP} & Author (A): 4,057 & Author-Paper: 19,645 & 334 & A: \textit{Database}, & APA\\
 & Paper (P): 14,328&  Paper-Conference: 14,328 &  & \textit{Data Mining}, & APCPA \\
 & Conference (C): 20 & Paper-Term-EP1: 88,420 & & \textit{Machine Learning}, & APTPA\\
 & Term (T)-EP1: 8,789 & Paper-Term-EP2: 85,810 & &\textit{Information Retrieval} & \\
  & Term (T)-EP2: 7,723  & & & & \\
 \midrule
 \multirow{4}{*}{IMDB-EP1} & Movie (M): 4,275 & Movie-Actor: 12,838 & 6,344 & M: \textit{Action}, & MAM\\
 & Director (P): 2,082&  Movie-Director: 4,280 &  &  \textit{Comedy}, & MDM \\
 & Actor (A): 5,431 & Movie-Keyword: 20,529  & & \textit{Drama} & MKM\\
 & Keyword (K): 7,313  & & & & \\
 \midrule
 \multirow{3}{*}{IMDB-EP2} & Movie (M): 4,278 & Movie-Actor: 12,838 & 1,232 & M: \textit{Action}, & MDM, MAM\\
 & Director (P): 2,081 &   Movie-Director: 4,278 &  & \textit{Comedy}, & DMD, DMAMD \\
 & Actor (A): 5,257 &  & & \textit{Drama} & AMA, AMDMA\\
 \midrule
 \multirow{3}{*}{Last.fm} & User (U): 1,892 & User-User: 12,717 & - &  & UU, UAU\\
 & Artist (A): 17,632&   User-Artist: 92,834 & identity & - &UATAU, AUA \\
 & Tag (T): 1,088 &  Artist-Tag: 23,253 & features &  & AUUA, ATA\\
 
\bottomrule
\end{tabular}
}
\end{table}
\addtolength{\tabcolsep}{-2pt}

\subsection{Tasks}
The node-related tasks for evaluation are node classification, node clustering, and link prediction.
    \begin{itemize}
    \item \textbf{Node classification}. In the node classification task, the unsupervised learned representations are fed to a downstream classifier, while the supervised methods output the classification result as end-to-end models. The classification is repeated 10 times, and average Macro-F1 and Micro-F1 scores are reported.
    
    In EP1, the training set sizes (i.e., given labels for training) is 20\% or 80\% of full datasets, and  the validation set and test set sizes are fixed at 10\% of full datasets. The downstream classifier is logistic regression classifier which is optimized using the Adam~\cite{kingma2014adam} with a learning rate  of~0.001 and with an early stopping on the validation accuracy.
    
    In EP2, the training, validation, and testing sets are of 400 (9.35\%), 400 (9.35\%), and 3478 (81.30\%) nodes respectively in IMDB and 400 (9.86\%), 400 (9.86\%), and 3257 (80.28\%) nodes respectively in DBLP. The  downstream classifier is a linear SVM classifier with default parameters, which only uses the test node representations  (i.e., 3478
nodes for IMDb and 3257 nodes for DBLP). For training, the test nodes are re-divided into training and testing proportions, namely 20\%, 40\%, 60\% and 80\%.
    \end{itemize}
    
    \begin{itemize}
    \item \textbf{Node clustering}. In the node clustering task, the $K$-Means algorithm is used to conduct the clustering based on the learned representations. The number of clusters $K$ is set as to the number of target node classes. Unsupervised methods do not use any label during learning representations, while supervised methods do. The clustering process is repeated  10 times and the average normalized mutual information (NMI) and adjusted rand index (ARI) evaluation metrics are reported (computed for all nodes in EP1 and computed for the test nodes in EP2). 
     \end{itemize}
     
     \begin{itemize}
    \item \textbf{Link prediction}. In link prediction, the goal is to correctly predict whether two nodes connect or do not connect. Usually, some positive and negative edges of the original graph are masked and used to test the predictions. Given two node representations $\mathbf{z}_u$ and $\mathbf{z}_v$ generated by the trained model, the probability that $u$ and $v$ link together is calculated as $p(u,v) = \sigma( {\mathbf{z}_u}^T \mathbf{z}_v)$,  where $\sigma(\cdot)$ is the sigmoid function. The probability should be close to 1 for positive edges and close to 0 for negative edges. The evaluation is performed by reporting the mean area under the ROC curve (AUC) and average precision (AP) scores after 5 runs.
    
    For EP1, we mask 50\% (45\% for testing and 5\% for validation) of edges (and we obtain an equal number of negative edges) of each meta-path and use the computed representations to predict them. Heterogeneous methods use the same representation for all meta-path link predictions, while homogeneous use the per-meta-path representation for each meta-path.
    
    For EP2 and Last.fm, the graph is divided into batches with size of 64. To make the task more challenging, we train the model on the subgraphs of the obtained batches, and evaluate to an equal number of completely unseen subgraphs of other batches.  The number of batches experimented is 1, 10, 50. 
     \end{itemize}
     
    \begin{itemize}
    \item \textbf{Loss augmentation with HeMI}. We also evaluate the effectiveness of HeMI by employing it as an additional loss function to end-to-end trained models. The final representations of the models are obtained by $\mathbf{h}_v = \mathbf{W}_{\text{task}} \mathbf{z}_v$, where $\mathbf{W}_{\text{task}}$ are learnable weights with proper dimensions. Then, $\mathbf{h}_v$ is optimized based on a task specific loss. 
    
    In semi-supervised node classification, we minimize the cross-entropy over all labeled nodes between the ground-truth labels and the predicted labels as 
    \begin{equation}
        \mathcal{L}_{\text{nc}} = - \sum_{v \in \mathcal{V}_Y} \mathbf{Y}_v \ln \mathbf{H}_v,
    \end{equation}
    where $\mathcal{V}_Y$ is the set of nodes that have labels,  and $\mathbf{Y}_v$ and $\mathbf{H}_v$  are the labels and the predicted label probabilities of the labeled nodes, respectively.
    
    In link prediction, we  minimize the binary cross-entropy loss function between positive and negative edges as
    \begin{equation}
        \mathcal{L}_{\text{lp}} = - \sum_{ \{(v,u)\}} \log \sigma( {\mathbf{h}_v}^T \mathbf{h}_u) - \sum_{\{(v,u')\}} \log \sigma( -{\mathbf{h}_v}^T \mathbf{h}_{u'}),
    \end{equation}
    where  $\sigma(\cdot)$ is the sigmoid function, $\{(v,u)\}$ is the set of observed (positive) node pairs, $\{(v,u')\}$ is the set of negative node pairs sampled from all unobserved node pairs (the complement of $\{(v,u)\}$).
    
    In both cases, HeMI is used to optimize the intermediate representations $\mathbf{z}_v$ based on Eq.~\eqref{eq:loss}.
    \end{itemize}

\subsection{Reproducibility}
For competing methods, we report their performance by reusing the reported results of~\cite{ren2019hdgi,fu2020magnn}. Whenever we need to implement a method by our own, we do it based on their original implementations and proposed configurations\footnote{\url{https://github.com/cmavro/Graph-InfoClust-GIC}}, and for a fair comparison, we use the same encoders for these methods as explained below. All competing approaches are described in Section~\ref{sec:rel}.

For HeMI, we set representations' dimensions to $d=256$ and the semantic attention dimensions to $d_m=16$, in all experiments; we only set $d=64$ in link prediction tasks. In EP1 experiments, we use  1-layer GCN encoders, and for IMDB dataset only, we use the same GCN encoder for all views. In EP2 experiments, we use either a neighbor-based MAGNN encoder ($\text{MAGNN}_{\text{nb}}$) or a RotatE~\cite{sun2019rotate} MAGNN encoder ($\text{MAGNN}_{\text{rot}}$) with layers selected from \{1,2\} and we use a mean aggregator in Eq.~\eqref{eq:aggr}. HeMI is optimized with Adam, with a learning rate of 0.001, for epochs selected from \{1000,2000\} with an early stopping of 50 patience epochs on the training loss. The hyper-parameter $\lambda$, which controls the relative importance of each objective term, is selected from \{0, .25, .5, .75, 1\}, with a parameter study provided in the Supplementary Material. 

We implement HeMI by modifying HDGI original code\footnote{\url{https://github.com/YuxiangRen/Heterogeneous-Deep-Graph-Infomax}} (which is implemented with PyTorch~\cite{paszke2017automatic}) for EP1, and by modifying MAGNN original code\footnote{\url{https://github.com/cynricfu/MAGNN}} (which is implemented with PyTorch and Deep Graph Library~\cite{wang2019dgl}) for EP2. All experiments were performed on a Nvidia Geforce RTX-2070 GPU on a i5-8400 CPU and 32GB RAM machine.

\addtolength{\tabcolsep}{2pt}   
\begin{table}[t]
\centering
\caption{Node classification and clustering performance for EP1.}
\label{tb:hyp-cl}
\resizebox{0.7\textwidth}{!}{ 
\begin{threeparttable}
\begin{tabular}{lc|cc|cc|cc}
\toprule
 & Fusion/ & \multicolumn{2}{c}{ACM} & \multicolumn{2}{c}{DBLP} & \multicolumn{2}{c}{IMDB} \\
 & Consensus& Classif. & Clust. & Classif. & Clust. & Classif. & Clust.\\
  \midrule
 DGI+concat & No/No & 93.28 & 67.31 & 93.21 & \textbf{74.97} & 59.54 & 1.61  \\
 GIC+concat & No/No & 93.18 & 67.92 & \textbf{93.28} & 73.33 & 62.11 & 5.95  \\
 MVGRL & No/Yes & 92.19 & 66.75 & 92.40 & 72.79 & 60.37 & 2.07  \\
 HDGI &  Yes/No & 92.27 & 71.83 & 91.75 & 69.25 & 58.93 & 1.87 \\
 HGIC &  Yes/No & 93.58 & 69.32 & 92.57 & 69.80 & 57.45 & 8.87 \\
 DMGI &  Yes/Yes & 92.55 & 67.48 & 80.96 & 56.07 & 58.88 & 2.18 \\
 \textbf{HeMI} & Yes/Yes  & \textbf{93.97} & \textbf{73.45}  & 93.06 & 69.98 & \textbf{63.65} & \textbf{12.23}\\

\bottomrule
\end{tabular}
\begin{tablenotes}
\item The scores reported are Micro-F1 with 20\% train nodes and NMI, for classification (Classif.) and clustering (Clust.), respectively. 
\end{tablenotes}

\end{threeparttable}
}
\end{table}
\addtolength{\tabcolsep}{-2pt}

\addtolength{\tabcolsep}{2pt}   
\begin{table}[t]
\centering
\caption{Link prediction performance for EP1.}
\label{tb:hyp-lp}
\resizebox{\textwidth}{!}{ 
\begin{threeparttable}
\begin{tabular}{lc|ccc|cccc}
\toprule
 &  & \multicolumn{3}{c}{ACM} &  \multicolumn{4}{c}{DBLP} \\
 & Fusion/ & PAP & PSP & avg. & APA & APTPA & APCPA &  avg. \\
 &Consensus & AUC/AP & AUC/AP & AUC/AP. & AUC/AP & AUC/AP &AUC/AP  &AUC/AP   \\
  \midrule
 DGI+concat & No/No & 63.26/62.34 & 51.27/54.38 & 57.26/58.36 & 73.79/72.10 & 61.30/62.61 & 72.00/76.50 & 69.03/70.41\\
 GIC+concat & No/No & \underline{84.38/79.63} & 44.53/56.76 & 64.46/68.20 & 76.25/74.99 & \underline{73.39/74.59} & 80.24/82.57 & 76.63/77.38\\
 MVGRL & No/Yes & 49.67/54.58 & 47.33/58.53 & 48.50/56.55 & 67.81/68.85 & 54.30/54.12 & 66.13/69.07 & 62.75/64.02 \\
 HDGI & Yes/No & 71.50/72.15 & 62.05/58.28 & 66.77/65.21 & 81.51/81.94 & 69.00/70.55 & 80.12/80.01 & 76.88/77.50 \\
 HGIC & Yes/No & 73.22/73.11 & 60.77/59.48 & 67.00/66.29 & 87.35/85.00 & 69.42/68.96 & 92.94/92.27 & 83.24/82.08 \\
 DMGI & Yes/Yes & 69.71/62.44 & 66.23/60.17 & 67.97/61.31 & 88.96/88.94 & 68.11/69.83 & 67.84/67.20  & 74.98/75.32 \\
 \textbf{HeMI} & Yes/Yes  & 74.69/74.64 & \underline{70.77/64.92} & \textbf{72.87/69.24} & \underline{89.83/87.54} & 69.60/67.37 & \underline{96.54/96.74} & \textbf{85.32/83.88 }  \\

\bottomrule
\end{tabular}
\begin{tablenotes}
\item We report scores for each meta-path graph (PAP and PSP in ACM, and APA, APTPA, and APCPA in DBLP) and overall scores of all meta-paths (avg.).
\end{tablenotes}

\end{threeparttable}
}
\end{table}
\addtolength{\tabcolsep}{-2pt}


\section{Results}

In this section, we wish to answer the following research questions (RQ):
\begin{itemize}
    \item \textit{RQ1}. Is meta-path fusion and consensus important when learning HG representations?
    \item \textit{RQ2}. How does the proposed method perform compared to other methods? 
    \item \textit{RQ3}. Can the proposed objective be used as an augmented loss function for better representation learning?
\end{itemize}

\subsection{Results using the first evaluation protocol (EP1)}\label{sec:hyp}
\textit{(RQ1\&RQ2)}. Table~\ref{tb:hyp-cl} shows that HeMI, which relies on both meta-path fusion and meta-path consensus, outperforms all other methods in ACM and IMDB. An exception is the DBLP dataset, since the labels (conference types) are directly obtained by the APCPA meta-path and thus, fusion does not convey additional information; here, DMGI performs the worst due to its strict consensus regularization. Moreover, for node classification, methods that simply concatenate the  per-meta-path representations  (DGI and GIC) can outperform their fusion-based counterparts (HDGI and HGIC) because the downstream classifier is powerful enough to extract hidden interactions, in the same way fusion does. However, this does not happen for node clustering since fusion-based methods compute more compact representations (with the exception of DBLP).

Table~\ref{tb:hyp-lp} shows that HeMI outperforms (avg. column) all other methods for link prediction tasks, indicating that is able to encode useful information from all meta-paths simultaneously. Note that fusion-based methods do not perform as well as they appear to overestimate information from specific meta-paths, e.g., HGIC and HDGI overestimate PAP in ACM. Although, non-fusion methods, like GIC, outperform other methods on some meta-paths, e.g., on PAP in ACM and APTPA in DBLP, they cannot generalize as well for all meta-paths, since information exchange is necessary. Finally, other consensus-based methods either lose the per-meta-path useful information for link prediction (MVGRL) or are not as powerful as HeMI (DMGI).

We also release full results for node classification, node  clustering, and link prediction in the Supplementary Material.

\addtolength{\tabcolsep}{2pt}    
\begin{table}[t]
\small
\centering
\caption{Node classification results for EP2.}
\label{tb:classif2}
\resizebox{\textwidth}{!}{  
\begin{tabular}{ccc|cccc|ccccccc}
\toprule
Dataset & Metric& Train&  \multicolumn{4}{c|}{Supervised} & \multicolumn{7}{c}{Unsupervised}\\
& & & GCN & GAT & HAN & MAGNN & LINE & N2V & ESim & M2V & HERec & HGIC & \textbf{HeMI}\\
\midrule
\multirow{8}{*}{IMDb} & \multirow{4}{*}{Macro-F1}& 20\% & 52.73 & 53.64 &56.19 & 59.35 & 44.04 & 49.00 &  48.37 & 46.05 & 45.61 & 57.93 & \underline{\textbf{59.40}}$\pm$ 0.59  \\
& & 40\% & 53.67 &55.50 &56.15& 60.27 & 45.45 &50.63& 50.09 &47.57 &46.80 &59.06  & \underline{\textbf{60.72}}$\pm$ 0.92\\
& & 60\% & 54.24& 56.46& 57.29& 60.66 & 47.09& 51.65& 51.45& 48.17 &46.84 & 59.53 & \underline{\textbf{61.51}}$\pm$ 0.90 \\
& & 80\% & 54.77& 57.43 &58.51 &61.44 & 47.49& 51.49& 51.37& 49.99 &47.73 & 59.57 & \underline{\textbf{62.87}}$\pm$ 1.31\\
\cline{2-14}
& \multirow{4}{*}{Micro-F1}& 20\% & 52.80& 53.64 &56.32& \underline{59.60} & 45.21& 49.94 &49.32& 47.22& 46.23 & 58.29 & \textbf{59.47}$\pm$ 0.58\\
& & 40\% & 53.76 &55.56 &57.32 & 60.50 & 46.92& 51.77& 51.21 &48.17& 47.89 & 59.32 & \underline{\textbf{60.76}}$\pm$ 0.87 \\
& & 60\% & 54.23& 56.47& 58.42& 60.88 & 48.35& 52.79& 52.53& 49.87& 48.19 & 59.81 & \underline{\textbf{61.51}}$\pm$ 0.92 \\
& & 80\% & 54.63& 57.40 &59.24 &61.53 & 48.98& 52.72 &52.54 &50.50& 49.11  & 59.84 & \underline{\textbf{62.88}}$\pm$ 1.29 \\
\midrule
\multirow{8}{*}{DBLP} & \multirow{4}{*}{Macro-F1}& 20\% & 88.00& 91.05& 91.69& 93.13 &87.16 &86.70 &90.68 &88.47& 90.82 & 92.66 & \textbf{93.50}$\pm$ 0.44  \\
& & 40\% & 89.00 &91.24& 91.96 &93.23 & 88.85& 88.07& 91.61& 89.91& 91.44&   92.85 & \textbf{93.67}$\pm$ 0.38\\
& & 60\% & 89.43& 91.42 &92.14 &93.57& 88.93& 88.69 &91.84& 90.50 &92.08 & 92.96 & \underline{\textbf{93.80}}$\pm$ 0.46 \\
& & 80\% & 89.98 &91.73 &92.50 &94.10 & 89.51 &88.93& 92.27 &90.86& 92.25 &  93.02 & \underline{\textbf{94.13}}$\pm$ 0.73\\
\cline{2-14}
& \multirow{4}{*}{Micro-F1}& 20\% & 88.51& 91.61& 92.33& 93.61 & 87.68& 87.21& 91.21& 89.02& 91.49& 93.33  & \underline{\textbf{94.01}}$\pm$ 0.42\\
& & 40\% & 89.22 &91.77 &92.57& 93.68 & 89.25& 88.51& 92.05& 90.36& 92.05 & 93.39 & \underline{\textbf{94.17}}$\pm$ 0.36 \\
& & 60\% & 89.57 &91.97& 92.72& 93.99 & 89.34& 89.09& 92.28& 90.94 &92.66 & 93.43 & \underline{\textbf{94.30}}$\pm$ 0.41 \\
& & 80\% & 90.33& 92.24 &93.23 &94.47 & 89.96 &89.37 &92.68 &91.31 &92.78  &  93.50& \underline{\textbf{94.59}}$\pm$ 0.67 \\




\bottomrule
\end{tabular}
}
\end{table}
\addtolength{\tabcolsep}{-2pt}

\addtolength{\tabcolsep}{2pt}   
\begin{table}[t]
\small
\centering
\caption{Node clustering results for EP2.}
\label{tb:clust2}
\resizebox{0.8\textwidth}{!}{ 
\begin{tabular}{cc|cccc|ccccccc}
\toprule
Dataset & Metric & \multicolumn{4}{c|}{Supervised} & \multicolumn{7}{c}{Unsupervised}\\
& & GCN & GAT & HAN & MAGNN & LINE & N2V & ESim & M2V & HERec & HGIC & \textbf{HeMI}\\
\midrule
\multirow{2}{*}{IMDb}& NMI &7.46 &7.84 &10.79 &\underline{15.58}& 1.13& 5.22& 1.07& 0.89 &0.39 & 7.2 & \textbf{9.74}\\
&ARI &7.69& 8.87 &11.11 &\underline{16.74} &1.20 &6.02& 1.01& 0.22& 0.11& 7.3 & \textbf{10.04}\\
\midrule
\multirow{2}{*}{DBLP}& NMI &73.45& 70.73& 77.49 &\underline{80.81} & 71.02 &77.01& 68.33& 74.18& 69.03& 70.51 & \textbf{77.18}\\
&ARI &77.50 &76.04 &82.95& \underline{85.54}&76.52 &81.37& 72.22 &78.11 &72.45& 76.19 & \textbf{81.71}\\
\bottomrule
\end{tabular}
}
\end{table}
\addtolength{\tabcolsep}{-2pt}

\subsection{Results using the second evaluation protocol (EP2)}

\textit{(RQ2)}. Tables~\ref{tb:classif2}~and~\ref{tb:clust2} show the node classification and clustering performance, respectively, for EP2. As it can be seen, HeMI outperforms both semi-supervised and unsupervised methods for node classification. Specifically,  HeMI outperforms the best performing unsupervised method by up to 3.3\% in IMDb and by up to 1.1\% in DBLP dataset. HeMI achieves slightly better than MAGNN (about 1\%) and outperforms other semi-supervised methods by more than 4\% and 1.3\% in IMDb and DBLP, respectively. For node clustering, HeMI performs better than other unsupervised HGNN/GNN methods in both datasets. However, it is not able to outperform semi-supervised methods HAN and MAGNN since they have already seen the labels during training. Here, we only report HGIC from methods of Section~\ref{sec:hyp}, since we found out that it performs the best with a MAGNN encoder and is a generalized version of HDGI.

\addtolength{\tabcolsep}{2pt}   
\begin{table}[t]
\caption{Link prediction and HeMI as objectives for Last.fm.}
\label{tb:hemilp}
\small
\centering
\resizebox{0.7\textwidth}{!}{ 
\begin{tabular}{l|cc|cc|cc}
\toprule
Train/Test & \multicolumn{2}{c}{1 batch} & \multicolumn{2}{c}{10 batches} & \multicolumn{2}{c}{50 batches}\\
 Metric & AUC & AP & AUC & AP & AUC & AP \\
 \midrule
  \textbf{Supervision} & & & & & & \\
Link Prediction & 65.65 & 56.50 & 67.95 & 56.85 & 87.46 & 83.74 \\
Link Prediction + HeMI ($\lambda=0$) & \textbf{78.56} & \textbf{66.96} & 71.49 & 60.12 & 88.28 & 83.80 \\
Link Prediction + HeMI ($\lambda=1$) & 73.17 & 63.92 & \textbf{75.52} & \textbf{65.83} & \textbf{94.37} & \textbf{95.16} \\
\bottomrule
\end{tabular}
}
\end{table}
\addtolength{\tabcolsep}{-2pt}

\addtolength{\tabcolsep}{2pt}   
\begin{table}[t]
\caption{Node classification and HeMI as objectives for IMDB-EP2 with multiple  labels.}
\label{tb:heminc}
\small
\centering
\resizebox{0.7\textwidth}{!}{ 
\begin{threeparttable}
\begin{tabular}{l|cc|cc|cc}
\toprule
 Labels & \multicolumn{2}{c}{\textit{Genre}} & \multicolumn{2}{c}{\textit{Rating}} & \multicolumn{2}{c}{\textit{Budget}} \\
 \midrule
 & Classif. & Clust. & Classif. & Clust. & Classif. & Clust.\\
  \midrule
  \textbf{Supervision} & & & & & & \\
\textit{Genre} & 61.53/59.60 & 15.58 & 38.89/37.99 & 0.52 & 44.08/42.33 & 2.50\\
\textit{Genre}+HeMI & 62.62/\textbf{61.70} & \textbf{16.63} & 42.41/40.73 & 1.34 & 48.86/47.48 & 2.87\\
\cline{2-7}
\textit{Rating}& 47.42/44.88 & 1.09 & \textbf{44.58}/41.95 & 2.75 & 45.47/43.32 & 0.06\\
\textit{Rating}+HeMI& 49.43/45.81 & 1.17 & 43.14/\textbf{42.69} & \textbf{3.26} & 46.76/45.12 & 1.95\\

\cline{2-7}
\textit{Budget}& 51.03/47.58 & 2.80 & 38.62/37.04 & 1.82 & 47.27/45.90 & 2.97\\
\textit{Budget}+HeMI& 51.32/48.45 & 3.97 & 39.95/37.41 & 2.47 & 49.79/47.15 & 4.43\\
\cline{2-7}
HeMI& \textbf{62.88}/59.47 & 9.74 & 43.91/41.05 & 1.04 & \textbf{54.21}/\textbf{51.14} & \textbf{7.29}\\
\bottomrule
\end{tabular}
\begin{tablenotes}
\item The scores reported are averaged the Micro-F1 with 80\%/20\% train nodes and NMI, for classification (Classif.) and clustering (Clust.), respectively.
\end{tablenotes}
\end{threeparttable}
}
\end{table}
\addtolength{\tabcolsep}{-2pt}

\subsection{Loss augmentation with HeMI}\label{sec:hess}

\textit{(RQ3)}. Table~\ref{tb:hemilp} shows that HeMI benefits the MAGNN encoder when used together for link prediction. Specifically, when using 1 batch for training and 1 for testing, the fine-grain consensus ($\lambda=0$) leads to superior performance by more than 10\% compared to only using the link prediction loss. Here, the coarse-grain consensus ($\lambda=1$) does not work as well, since there are few nodes in the batch. By increasing the batches, the coarse-grain consensus' power is also increased and outperforms the link-prediction-only supervision by 8\%, in average. 

Table~\ref{tb:heminc} shows that HeMI preserves useful information that can be used for various node classification tasks, compared to a supervised MAGNN encoder. When using \textit{Genre} labels with HeMI self-supervision, it outperforms the label-only supervision by more than 2.5\%, in average. \textit{Genre} labels depend more on the plot of a movie (node attributes) and HeMI gives extra importance on the features by using fake attributes. 
\textit{Budget} label depends usually on the cast of a movie, e.g., director and actor, and consequently on the information MDM and MAM meta-paths provide. HeMI is designed to preserve this information and thus, leads to superior performance by more than 1\%, in average.  \textit{Rating} labels are more challenging to predict, and thus, using labels during training facilitates the learning. Finally, HeMI as a self-supervision alone outperforms its semi-supervised counterpart for all labels and tasks, which indicates that it preserves all the information useful for the classification tasks.

\section{Conclusion}

In this paper, we have proposed HeMI for learning representations from heterogeneous graphs. HeMI relies on meta-path information fusion and consensus to optimize the representations. Experiments show that HeMI outperforms competing approaches in various node-related tasks, such as node classification, node clustering, and link prediction. 

\subsubsection*{Acknowledgements}
This work was supported in part by NSF (1447788, 1704074, 1757916, 1834251), Army Research Office (W911NF1810344), Intel Corp, and the Digital Technology Center at the University of Minnesota. Access to research and computing facilities was provided by the Digital Technology Center and the Minnesota Supercomputing Institute.


\bibliographystyle{splncs04}
\bibliography{HetMI}


\clearpage
\section{Supplementary Material}

\textbf{Dataset descriptions}.

\begin{itemize}
    \item \textbf{IMDB}. The IMDB dataset contains 4,780 movies (M), 5,841 actors (A), 2,269 directors (D) and 7,313 keyword (K). The movies are divided into three classes (Action, Comedy, Drama) according to their genre. In EP1, the movie features are composed of \{color, title, language, keywords, country, rating, year\} with a TF-IDF encoding. The meta-paths chosen are MAM, MDM, and MKM. On the other hand, in EP2, the movie features correspond to elements of a bag-of-words represented of plots and the meta-path set is \{MAM, MDM, DMD, DMAMD, AMA, AMDMA\}. 
    
    In order to show that our method's learned representations are useful for various tasks, we manually extract two additional set of classes from the raw data. The first set of classes corresponds to movie ratings (rating $<$ 6, 6 $\leq$ rating $<$ 7, 7 $\leq$ rating) and the second to movie budgets (budget $<$ \$10M, \$10M $\leq$ budget $<$ \$50M, \$50M $\leq$ budget). We choose the specific boundary values of each class label, so that classes are equally distributed among the movie instances. 
    
    \item \textbf{DBLP}. The DBLP dataset contains 14328 papers (P), 4057 authors (A), 20 conferences (C), 8789 terms (T). The authors are divided into four areas: database, data mining, machine learning, information retrieval. Also, each author’s research area is labeled according to the conferences they submitted. For both EP1 and EP2, author features are the elements of a bag-of-words represented of keywords and the meta-path set employed the meta-path is \{APA, APCPA, APTPA\}.

    \item \textbf{ACM}. The ACM dataset is constructed by papers published in KDD, SIGMOD, SIGCOMM, MobiCOMM, and VLDB and are divided into three classes (Database, Wireless Communication, Data Mining). The heterogeneous graph comprises 3025 papers (P), 5835 authors (A) and 56 subjects (S). Paper features correspond to elements of a bag-of-words represented of keywords and the meta-path set is {PAP, PSP}. Here, the papers are labeled according to the conference they published and this dataset is used in EP1 only.

    \item \textbf{Last.fm}. The dataset consists of 1892 users (U), 17632 artists (A), and 1088 artist tags (T). Last.fm is used for the link prediction task, no labels or features are included, it is used in EP2 only, and the selected meta-paths are \{UU, UAU, UATAU, AUA, AUUA, ATA\}.
    
\end{itemize}

\addtolength{\tabcolsep}{2pt}    
\begin{table}[t]
\small
\centering
\caption{Node classification results for EP1.}
\label{tb:classif1}
\resizebox{\textwidth}{!}{  
\begin{threeparttable}
\begin{tabular}{ccc|rrrr|rrrrrrc}
\toprule
Dataset & Train & Metric& \multicolumn{4}{c|}{Supervised} & \multicolumn{6}{c}{Unsupervised}\\
 & & & GCN & RGCN & GAT & HAN & RAW & M2V & DW & DGI & HDGI & (H)GIC & \textbf{HeMI} \\
 \midrule
 \multirow{4}{*}{ACM} &\multirow{2}{*}{20\%} &Micro-F1 & 0.9250&  0.5766 & 0.9178 &0.9267& 0.8590& 0.6125& 0.8785& 0.9104 & 0.9227& 0.9358 &\textbf{0.9397}$\pm$ .0111\\
 & &Macro-F1&0.9250&  0.5766 & 0.9178 &0.9267& 0.8590& 0.6125& 0.8785& 0.9104 & 0.9227& 0.9361 & \textbf{0.9407}$\pm$ .0109\\
  \cline{2-14}
 &\multirow{2}{*}{80\%} &Micro-F1 & 0.9317 &0.5939& 0.9250& 0.9400& 0.8820 &0.6378& 0.8965& 0.9175& 0.9379& 0.9378 &\textbf{0.9447}$\pm$ .0101\\
 & &Macro-F1&0.9250&  0.5766 & 0.9178 &0.9267& 0.8590& 0.6125& 0.8785& 0.9104 & 0.9227& 0.9393 &\textbf{0.9455}$\pm$ .0107\\
 \midrule
 \multirow{4}{*}{DBLP} & \multirow{2}{*}{20\%} &Micro-F1& 0.8192 & 0.1932 & 0.8244 & 0.8992& 0.7552 & 0.6985& 0.7163 &0.8975&0.9175& \textbf{0.9328} & 0.9306$\pm$ .0122\\
 & &Macro-F1& 0.8128 &0.2132 &0.8148& 0.8923& 0.7473 &0.6874& 0.7063 & 0.8921&0.9094& \textbf{0.9271} & 0.9265$\pm$ .0126\\
  \cline{2-14}
  & \multirow{2}{*}{80\%} &Micro-F1& 0.8383 &0.2175 &0.8540& 0.9100& 0.8325 &0.8211& 0.7860 &0.9150 & 0.9226& \textbf{0.9333} &0.9321$\pm$ .0164\\
 & &Macro-F1& 0.8308& 0.2212& 0.8476 &0.9055& 0.8152 &0.8014& 0.7799& 0.9052 &0.9106& \textbf{0.9282} &0.9277$\pm$ .0171\\

 \midrule
 \multirow{4}{*}{IMDB} & \multirow{2}{*}{20\%} &Micro-F1& 0.5931& 0.4350& 0.5985& 0.6077&0.5112&0.3985& 0.5262& 0.5728& 0.5893& 0.6211 & \textbf{0.6365}$\pm$ .0244\\
 & &Macro-F1& 0.5869& 0.4468& 0.5944& 0.6027&0.5107& 0.4012 & 0.5293 &0.5690& 0.5914 & 0.6222 & \textbf{0.6367}$\pm$ .0254\\
 \cline{2-14}
& \multirow{2}{*}{80\%} &Micro-F1& 0.6467& 0.4476& 0.6540& 0.6600&0.5900& 0.4203& 0.6017& 0.6003& 0.6592 & 0.6642 & \textbf{0.6808}$\pm$ .0229\\
 & &Macro-F1& 0.6457& 0.4527& 0.6550& 0.6586&0.5884 &0.4119 & 0.6049 &0.5950& 0.6646 & 0.6655 &\textbf{0.6794}$\pm$ .0231\\
\bottomrule
\end{tabular}
\begin{tablenotes}
\item (H)GIC: We report the best result between GIC+concat and HGIC. 
\item DGI: We report the best result of using DGI on each single meta-path.
\item DW: We report the best result of DeepWalk on each single meta-path with concatenated features.
\end{tablenotes}

\end{threeparttable}
}
\end{table}
\addtolength{\tabcolsep}{-2pt}

\addtolength{\tabcolsep}{2pt}   
\begin{table}[t]
\small
\centering
\caption{Node clustering results for EP1.}
\label{tb:clust1}
\resizebox{0.8\textwidth}{!}{ 
\begin{threeparttable}
\begin{tabular}{lcccccc}
\toprule
 & \multicolumn{2}{c}{ACM} & \multicolumn{2}{c}{DBLP} &\multicolumn{2}{c}{IMDB}\\
 & NMI & ARI & NMI & ARI & NMI & ARI\\
 \midrule
 DeepWalk& 25.47 &18.24 &7.40 &5.30 &1.23& 1.22\\
Raw Feature &32.62& 30.99 &11.21& 6.98& 1.06 &1.17\\
DeepWalk+Features& 32.54& 31.20 &11.98 &6.99 &1.23 &1.22\\
Metapath2vec& 27.59 &24.57 &34.30 &37.54 &1.15 &1.51\\
DGI+concat & 67.31 & 70.12 & \textbf{74.97} & \textbf{79.36} & 1.61 & 3.62\\
HDGI& 71.83& 74.56 & 69.25& 74.21 & 1.87 & 3.70\\
(H)GIC & 69.32 & 72.86 & 73.33& 78.55 & 8.87 & 9.09\\
\textbf{HeMI} & \textbf{73.45} & \textbf{77.58} & 69.98 & 75.76 & \textbf{12.23} & \textbf{12.48}\\
 \bottomrule
\end{tabular}
\begin{tablenotes}
\item (H)GIC: We report the best result between GIC+concat and HGIC. 
\end{tablenotes}

\end{threeparttable}
}
\end{table}
\addtolength{\tabcolsep}{-2pt}

\addtolength{\tabcolsep}{2pt}   
\begin{table}[t]
\small
\centering
\caption{Link prediction performance for EP1.}
\label{tb:hyp-lp2}
\resizebox{0.8\textwidth}{!}{ 
\begin{tabular}{lc|cccc}
\toprule
 &  & \multicolumn{4}{c}{IMDB} \\
 & Fusion/ & MDM & MAM & MKM & avg.\\
 & Consensus & AUC/AP & AUC/AP & AUC/AP. & AUC/AP  \\
  \midrule
 DGI+concat & No/No & 68.91/67.18 & 51.63/47.80 & 73.85/73.32 & 64.79/62.77  \\
 GIC+concat & No/No & 74.22/72.11 & \underline{70.63/67.75} & 76.53/77.11 &  \textbf{73.79/72.32}\\
 MVGRL & No/Yes &  51.85/53.13 & 45.39/45.48 & 70.53/68.76 & 55.92/55.79  \\
 HDGI & Yes/No & 50.78/50.49 & 54.09/52.13 &  70.13/68.80 & 58.33/57.14  \\
 HGIC & Yes/No & 52.29/52.62 &  53.73/53.47 &  \underline{83.41/82.35} &  63.14/62.81 \\
 DMGI & Yes/Yes &\underline{86.98/82.55} &  56.39/55.20 &  63.58/60.49 &  68.98/66.08 \\
 \textbf{HeMI} & Yes/Yes  & 55.85/55.74 & 61.18/58.80 & 72.70/72.89 & 63.24/62.48   \\

\bottomrule
\end{tabular}
}
\end{table}
\addtolength{\tabcolsep}{-2pt}

\begin{figure}
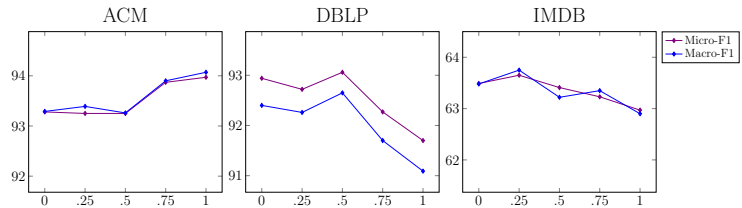

    \centering
    \includestandalone[width=0.8\textwidth]{figs/lambda}
    \caption{Ablation study on classification performance in EP1 with 20\% of train nodes (y-axis: scores in \%, x-axis: $\lambda $ values). } 
    \label{fig:lambda}
\end{figure}

\begin{figure}
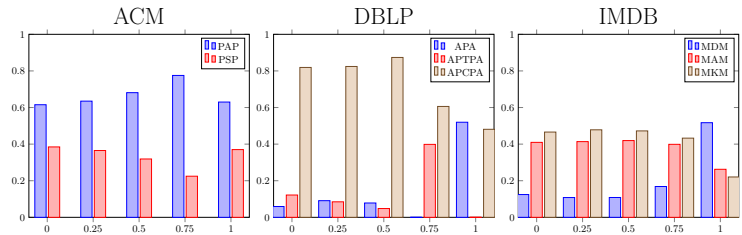

    \centering
    \includestandalone[width=0.8\textwidth]{figs/atts}
    \caption{Attention weights for each meta-path (y-axis: attention values, x-axis: $\lambda$ values). } 
    \label{fig:atts}
\end{figure}

\textbf{Comparison against other methods.}
Table~\ref{tb:classif1} shows HeMI's performance against other methods for node classification. In ACM, HeMI outperforms the best competing self-supervised method by more than 0.5\% and the best competing semi-supervised method by more than 1\%, in average.  In DBLP, the labels relate directly to the APCPA meta-path, so other meta-paths' information does not help for learning. Here, GIC+concat (denoted as (H)GIC), outperforms other heterogeneous methods (HeMI included). However, HeMI outperforms the best competing semi-supervised and self-supervised methods by up to 3\% and 2.5\%, respectively, in average. 
In IMDB, HeMI achieves significantly better results, outperforming the best competing approach by more than 1.5\%, in average.

Table~\ref{tb:clust1} shows HeMI's performance against other methods for node clustering. In ACM and IMDB, HeMI performs the best and the overall performance gain is more than 4\% compared to the best competing approach for both datasets. In DBLP, homogeneous-based methods, such as DGI+concat, perform the best, like in node classification. Finally, we would like to note here that using the same MI estimators for both meta-paths, and thus promoting additional consensus, achieved better performance than using different MI estimators, which perform about 0.5\% worse than the scores reported. 

Table~\ref{tb:hyp-lp2} shows the performance in the IMDB dataset for the link prediction task, where HeMI ranks 3rd. HeMI gives additional importance to the node features (through the negative examples of its loss function), and for IMDB, features are directly related to the MKM meta-path, which results into the overestimation of this meta-path. GIC is a powerful link prediction method and performs well in all meta-paths, while DMGI gains its power by performing very well on the MDM meta-path (although it neglects the other two meta-paths).

\textbf{Parameter Study}. Fig.~\ref{fig:lambda} shows a parameter study of the $\lambda$ regularization of HeMI's loss function. Different $\lambda$ values seem to differently affect the performance on each dataset. In order to explain the results, we also provide Fig.~\ref{fig:atts}, which shows the leraned importance of each meta-path.

In ACM, coarse-grain consensus works the best ($\lambda = 1$), although it computes similar weights for each meta-path with the fine-grain consensus. We believe this happens because the coarse-grain information helps to promote more useful semantics for the PAP and PSP meta-paths. In DBLP, the labels are directly related to the APCPA meta-path, so more weight needs to be given to this meta-path. For $\lambda=0.5$ (and values below), this happens and the best performance is achieved; while the coarse-grain consensus downgrades the performance since it under-estimates APCPA and completely neglects APA or APTPA. In IMDB, we use the same GNN encoder for all meta-paths and thus promote additional consensus. All $\lambda$ values compute almost identical weights for the meta-paths. An exception is for $\lambda = 1$, which performs the worst. 

Finally, we note that the attention weights computed by the supervised method HAN are: $\sim$0.7 PAP and $\sim$0.3 PSP for ACM, and $\sim$0.25 APA, $\sim$0.7  APCPA and $\sim$0.05 APTPA for DBLP. Our method computes very similar attention weights, although it does not use any labels, and we assume that these weight values are the ones that benefit the classification and clustering tasks.

\end{document}

%% file: figs/example.tex
\tikzset{every picture/.style={line width=0.75pt}} 

\begin{tikzpicture}[x=0.75pt,y=0.75pt,yscale=-1,xscale=1]

\draw  [dash pattern={on 0.84pt off 2.51pt}]  (643,115) -- (10,114) ;
\draw  [dash pattern={on 0.84pt off 2.51pt}]  (329,13) -- (330,102) ;
\draw [color={rgb, 255:red, 126; green, 211; blue, 33 }  ,draw opacity=1 ][line width=1.5]    (114,68) -- (148,68) ;
\draw [color={rgb, 255:red, 74; green, 144; blue, 226 }  ,draw opacity=1 ][line width=1.5]  [dash pattern={on 5.63pt off 4.5pt}]  (213,68) -- (233,68) -- (251,68) ;
\draw   (38.5,13) -- (48,22) -- (38.5,31) -- (29,22) -- cycle ;
\draw   (248,12) -- (265,12) -- (265,29) -- (248,29) -- cycle ;
\draw [color={rgb, 255:red, 208; green, 2; blue, 27 }  ,draw opacity=1 ][line width=1.5]    (4,69) .. controls (5.67,67.33) and (7.33,67.33) .. (9,69) .. controls (10.67,70.67) and (12.33,70.67) .. (14,69) .. controls (15.67,67.33) and (17.33,67.33) .. (19,69) -- (21,69) -- (21,69) .. controls (22.67,67.33) and (24.33,67.33) .. (26,69) .. controls (27.67,70.67) and (29.33,70.67) .. (31,69) .. controls (32.67,67.33) and (34.33,67.33) .. (36,69) -- (39,69) -- (39,69) ;
\draw   (143,19.5) .. controls (143,14.25) and (147.25,10) .. (152.5,10) .. controls (157.75,10) and (162,14.25) .. (162,19.5) .. controls (162,24.75) and (157.75,29) .. (152.5,29) .. controls (147.25,29) and (143,24.75) .. (143,19.5) -- cycle ;
\draw [color={rgb, 255:red, 208; green, 2; blue, 27 }  ,draw opacity=1 ][line width=1.5]    (416,24) .. controls (417.93,22.65) and (419.57,22.93) .. (420.93,24.86) .. controls (422.28,26.79) and (423.92,27.07) .. (425.85,25.72) .. controls (427.78,24.37) and (429.43,24.66) .. (430.78,26.59) .. controls (432.13,28.52) and (433.77,28.8) .. (435.7,27.45) .. controls (437.63,26.1) and (439.27,26.38) .. (440.63,28.31) .. controls (441.98,30.24) and (443.62,30.52) .. (445.55,29.17) .. controls (447.48,27.82) and (449.12,28.1) .. (450.48,30.03) .. controls (451.83,31.96) and (453.47,32.25) .. (455.4,30.9) .. controls (457.33,29.55) and (458.97,29.83) .. (460.33,31.76) .. controls (461.68,33.69) and (463.32,33.97) .. (465.25,32.62) .. controls (467.18,31.27) and (468.82,31.55) .. (470.18,33.48) .. controls (471.53,35.41) and (473.17,35.69) .. (475.1,34.34) .. controls (477.03,32.99) and (478.67,33.27) .. (480.03,35.2) .. controls (481.38,37.13) and (483.02,37.42) .. (484.95,36.07) .. controls (486.88,34.72) and (488.52,35) .. (489.88,36.93) .. controls (491.23,38.86) and (492.87,39.14) .. (494.8,37.79) -- (496,38) -- (496,38) ;
\draw [color={rgb, 255:red, 208; green, 2; blue, 27 }  ,draw opacity=1 ][line width=1.5]    (416,24) .. controls (417.37,22.08) and (419.01,21.81) .. (420.93,23.18) .. controls (422.85,24.55) and (424.49,24.28) .. (425.86,22.36) .. controls (427.23,20.44) and (428.88,20.16) .. (430.8,21.53) .. controls (432.72,22.9) and (434.36,22.63) .. (435.73,20.71) .. controls (437.1,18.79) and (438.74,18.52) .. (440.66,19.89) .. controls (442.58,21.26) and (444.22,20.99) .. (445.59,19.07) .. controls (446.96,17.15) and (448.6,16.88) .. (450.52,18.25) .. controls (452.44,19.62) and (454.09,19.34) .. (455.46,17.42) .. controls (456.83,15.5) and (458.47,15.23) .. (460.39,16.6) .. controls (462.31,17.97) and (463.95,17.7) .. (465.32,15.78) .. controls (466.69,13.86) and (468.33,13.59) .. (470.25,14.96) -- (473,14.5) -- (473,14.5) ;
\draw [color={rgb, 255:red, 74; green, 144; blue, 226 }  ,draw opacity=1 ][line width=1.5]  [dash pattern={on 5.63pt off 4.5pt}]  (492,14.5) -- (546,11) ;
\draw   (512,143.5) .. controls (512,138.13) and (516.39,133.78) .. (521.81,133.78) .. controls (527.23,133.78) and (531.63,138.13) .. (531.63,143.5) .. controls (531.63,148.87) and (527.23,153.22) .. (521.81,153.22) .. controls (516.39,153.22) and (512,148.87) .. (512,143.5) -- cycle ;
\draw   (575.37,130.72) .. controls (575.37,125.35) and (579.77,121) .. (585.19,121) .. controls (590.61,121) and (595,125.35) .. (595,130.72) .. controls (595,136.09) and (590.61,140.44) .. (585.19,140.44) .. controls (579.77,140.44) and (575.37,136.09) .. (575.37,130.72) -- cycle ;
\draw   (574.25,161.28) .. controls (574.25,155.91) and (578.64,151.56) .. (584.06,151.56) .. controls (589.48,151.56) and (593.88,155.91) .. (593.88,161.28) .. controls (593.88,166.65) and (589.48,171) .. (584.06,171) .. controls (578.64,171) and (574.25,166.65) .. (574.25,161.28) -- cycle ;
\draw [color={rgb, 255:red, 126; green, 211; blue, 33 }  ,draw opacity=1 ][line width=1.5]    (575.37,130.72) -- (531.63,143.5) ;
\draw [color={rgb, 255:red, 126; green, 211; blue, 33 }  ,draw opacity=1 ][line width=1.5]    (574.25,161.28) -- (531.63,143.5) ;
\draw   (406.5,15) -- (416,24) -- (406.5,33) -- (397,24) -- cycle ;
\draw   (473,14.5) .. controls (473,9.25) and (477.25,5) .. (482.5,5) .. controls (487.75,5) and (492,9.25) .. (492,14.5) .. controls (492,19.75) and (487.75,24) .. (482.5,24) .. controls (477.25,24) and (473,19.75) .. (473,14.5) -- cycle ;
\draw   (25,140) .. controls (25,134.75) and (29.25,130.5) .. (34.5,130.5) .. controls (39.75,130.5) and (44,134.75) .. (44,140) .. controls (44,145.25) and (39.75,149.5) .. (34.5,149.5) .. controls (29.25,149.5) and (25,145.25) .. (25,140) -- cycle ;
\draw   (405.5,54) -- (415,63) -- (405.5,72) -- (396,63) -- cycle ;
\draw   (455,63) .. controls (455,57.75) and (459.25,53.5) .. (464.5,53.5) .. controls (469.75,53.5) and (474,57.75) .. (474,63) .. controls (474,68.25) and (469.75,72.5) .. (464.5,72.5) .. controls (459.25,72.5) and (455,68.25) .. (455,63) -- cycle ;
\draw [color={rgb, 255:red, 208; green, 2; blue, 27 }  ,draw opacity=1 ][line width=1.5]    (415,63) .. controls (416.67,61.33) and (418.33,61.33) .. (420,63) .. controls (421.67,64.67) and (423.33,64.67) .. (425,63) .. controls (426.67,61.33) and (428.33,61.33) .. (430,63) .. controls (431.67,64.67) and (433.33,64.67) .. (435,63) .. controls (436.67,61.33) and (438.33,61.33) .. (440,63) .. controls (441.67,64.67) and (443.33,64.67) .. (445,63) .. controls (446.67,61.33) and (448.33,61.33) .. (450,63) .. controls (451.67,64.67) and (453.33,64.67) .. (455,63) -- (455,63) ;
\draw [color={rgb, 255:red, 126; green, 211; blue, 33 }  ,draw opacity=1 ][line width=1.5]    (482.5,24) -- (464.5,53.5) ;
\draw [color={rgb, 255:red, 126; green, 211; blue, 33 }  ,draw opacity=1 ][line width=1.5]    (498,44) -- (473,57) ;
\draw   (545,4) -- (562,4) -- (562,21) -- (545,21) -- cycle ;
\draw   (561,35) -- (578,35) -- (578,52) -- (561,52) -- cycle ;
\draw [color={rgb, 255:red, 74; green, 144; blue, 226 }  ,draw opacity=1 ][line width=1.5]  [dash pattern={on 5.63pt off 4.5pt}]  (492,14.5) -- (562,43) ;
\draw   (536,57) -- (553,57) -- (553,74) -- (536,74) -- cycle ;
\draw [color={rgb, 255:red, 74; green, 144; blue, 226 }  ,draw opacity=1 ][line width=1.5]  [dash pattern={on 5.63pt off 4.5pt}]  (514,44) -- (536,57) ;
\draw [color={rgb, 255:red, 74; green, 144; blue, 226 }  ,draw opacity=1 ][line width=1.5]  [dash pattern={on 5.63pt off 4.5pt}]  (474,63) -- (536,66) ;
\draw   (493,34.5) .. controls (493,29.25) and (497.25,25) .. (502.5,25) .. controls (507.75,25) and (512,29.25) .. (512,34.5) .. controls (512,39.75) and (507.75,44) .. (502.5,44) .. controls (497.25,44) and (493,39.75) .. (493,34.5) -- cycle ;
\draw   (85,141) .. controls (85,135.75) and (89.25,131.5) .. (94.5,131.5) .. controls (99.75,131.5) and (104,135.75) .. (104,141) .. controls (104,146.25) and (99.75,150.5) .. (94.5,150.5) .. controls (89.25,150.5) and (85,146.25) .. (85,141) -- cycle ;
\draw   (165,142) .. controls (165,136.75) and (169.25,132.5) .. (174.5,132.5) .. controls (179.75,132.5) and (184,136.75) .. (184,142) .. controls (184,147.25) and (179.75,151.5) .. (174.5,151.5) .. controls (169.25,151.5) and (165,147.25) .. (165,142) -- cycle ;
\draw   (252,142) .. controls (252,136.75) and (256.25,132.5) .. (261.5,132.5) .. controls (266.75,132.5) and (271,136.75) .. (271,142) .. controls (271,147.25) and (266.75,151.5) .. (261.5,151.5) .. controls (256.25,151.5) and (252,147.25) .. (252,142) -- cycle ;
\draw   (220.5,132) -- (230,141) -- (220.5,150) -- (211,141) -- cycle ;
\draw [color={rgb, 255:red, 126; green, 211; blue, 33 }  ,draw opacity=1 ][line width=1.5]    (85,141) -- (44,140) ;
\draw [color={rgb, 255:red, 208; green, 2; blue, 27 }  ,draw opacity=1 ][line width=1.5]    (184,142) .. controls (185.61,140.27) and (187.27,140.21) .. (189,141.81) .. controls (190.73,143.42) and (192.39,143.36) .. (193.99,141.63) .. controls (195.6,139.9) and (197.26,139.84) .. (198.99,141.44) .. controls (200.72,143.05) and (202.38,142.99) .. (203.99,141.26) .. controls (205.59,139.53) and (207.25,139.47) .. (208.98,141.07) -- (211,141) -- (211,141) ;
\draw [color={rgb, 255:red, 208; green, 2; blue, 27 }  ,draw opacity=1 ][line width=1.5]    (230,141) .. controls (231.74,139.41) and (233.4,139.49) .. (234.99,141.23) .. controls (236.58,142.97) and (238.25,143.04) .. (239.99,141.45) .. controls (241.73,139.86) and (243.39,139.94) .. (244.98,141.68) .. controls (246.57,143.42) and (248.24,143.5) .. (249.98,141.91) -- (252,142) -- (252,142) ;
\draw   (340,141) .. controls (340,135.75) and (344.25,131.5) .. (349.5,131.5) .. controls (354.75,131.5) and (359,135.75) .. (359,141) .. controls (359,146.25) and (354.75,150.5) .. (349.5,150.5) .. controls (344.25,150.5) and (340,146.25) .. (340,141) -- cycle ;
\draw   (427,141) .. controls (427,135.75) and (431.25,131.5) .. (436.5,131.5) .. controls (441.75,131.5) and (446,135.75) .. (446,141) .. controls (446,146.25) and (441.75,150.5) .. (436.5,150.5) .. controls (431.25,150.5) and (427,146.25) .. (427,141) -- cycle ;
\draw   (386,132) -- (403,132) -- (403,149) -- (386,149) -- cycle ;
\draw [color={rgb, 255:red, 74; green, 144; blue, 226 }  ,draw opacity=1 ][line width=1.5]  [dash pattern={on 5.63pt off 4.5pt}]  (359,141) -- (387,141) ;
\draw [color={rgb, 255:red, 74; green, 144; blue, 226 }  ,draw opacity=1 ][line width=1.5]  [dash pattern={on 5.63pt off 4.5pt}]  (405,141) -- (427,141) ;
\draw  [dash pattern={on 0.84pt off 2.51pt}]  (484,118) -- (483,227) ;

\draw (10.9,30.43) node [anchor=north west][inner sep=0.75pt]   [align=left] {Keyword};
\draw (133.16,33.92) node [anchor=north west][inner sep=0.75pt]   [align=left] {Paper};
\draw (234.38,34.9) node [anchor=north west][inner sep=0.75pt]   [align=left] {Author};
\draw (46,90) node [anchor=north west][inner sep=0.75pt]  [font=\large] [align=left] {(a) Node and edge types};
\draw (153,58) node [anchor=north west][inner sep=0.75pt]   [align=left] {cites};
\draw (258,60) node [anchor=north west][inner sep=0.75pt]   [align=left] {writes};
\draw (403,90) node [anchor=north west][inner sep=0.75pt]  [font=\large] [align=left] {(b) Heterogeneous graph};
\draw (171,195) node [anchor=north west][inner sep=0.75pt]  [font=\large] [align=left] {(c) Meta-paths};
\draw (489,178) node [anchor=north west][inner sep=0.75pt]  [font=\large] [align=left] {(d) Meta-path-based\\ neighbors};
\draw (43,59) node [anchor=north west][inner sep=0.75pt]   [align=left] {contains};
\draw (19,153) node [anchor=north west][inner sep=0.75pt]   [align=left] {Paper-Paper};
\draw (19,170) node [anchor=north west][inner sep=0.75pt]   [align=left] {(PP)};
\draw (151,157) node [anchor=north west][inner sep=0.75pt]   [align=left] {Paper-Keyword-Paper};
\draw (151,170) node [anchor=north west][inner sep=0.75pt]   [align=left] {(PKP)};
\draw (326,156) node [anchor=north west][inner sep=0.75pt]   [align=left] {Paper-Author-Paper};
\draw (326,170) node [anchor=north west][inner sep=0.75pt]   [align=left] {(PAP)};

\end{tikzpicture}

%% file: figs/framework.tex
\tikzset{every picture/.style={line width=0.75pt}} 

\begin{tikzpicture}[x=0.75pt,y=0.75pt,yscale=-1,xscale=1]

\draw  [fill={rgb, 255:red, 139; green, 87; blue, 42 }  ,fill opacity=0.4 ] (103.52,33.5) .. controls (103.52,29.72) and (106.78,26.66) .. (110.79,26.66) .. controls (114.8,26.66) and (118.06,29.72) .. (118.06,33.5) .. controls (118.06,37.27) and (114.8,40.33) .. (110.79,40.33) .. controls (106.78,40.33) and (103.52,37.27) .. (103.52,33.5) -- cycle ;
\draw  [fill={rgb, 255:red, 126; green, 211; blue, 33 }  ,fill opacity=0.4 ] (86.42,62.43) .. controls (86.42,58.66) and (89.67,55.6) .. (93.69,55.6) .. controls (97.7,55.6) and (100.96,58.66) .. (100.96,62.43) .. controls (100.96,66.2) and (97.7,69.26) .. (93.69,69.26) .. controls (89.67,69.26) and (86.42,66.2) .. (86.42,62.43) -- cycle ;
\draw  [fill={rgb, 255:red, 126; green, 211; blue, 33 }  ,fill opacity=0.4 ] (105.23,78.5) .. controls (105.23,71.62) and (111.17,66.05) .. (118.49,66.05) .. controls (125.81,66.05) and (131.74,71.62) .. (131.74,78.5) .. controls (131.74,85.39) and (125.81,90.96) .. (118.49,90.96) .. controls (111.17,90.96) and (105.23,85.39) .. (105.23,78.5) -- cycle ;
\draw  [fill={rgb, 255:red, 139; green, 87; blue, 42 }  ,fill opacity=0.4 ] (135.16,55.2) .. controls (135.16,51.42) and (138.41,48.36) .. (142.43,48.36) .. controls (146.44,48.36) and (149.7,51.42) .. (149.7,55.2) .. controls (149.7,58.97) and (146.44,62.03) .. (142.43,62.03) .. controls (138.41,62.03) and (135.16,58.97) .. (135.16,55.2) -- cycle ;
\draw  [fill={rgb, 255:red, 208; green, 2; blue, 27 }  ,fill opacity=0.4 ] (96.68,105.03) .. controls (96.68,101.25) and (99.94,98.2) .. (103.95,98.2) .. controls (107.96,98.2) and (111.22,101.25) .. (111.22,105.03) .. controls (111.22,108.8) and (107.96,111.86) .. (103.95,111.86) .. controls (99.94,111.86) and (96.68,108.8) .. (96.68,105.03) -- cycle ;
\draw  [fill={rgb, 255:red, 74; green, 144; blue, 226 }  ,fill opacity=0.4 ] (131.74,98.6) .. controls (131.74,94.83) and (134.99,91.77) .. (139.01,91.77) .. controls (143.02,91.77) and (146.28,94.83) .. (146.28,98.6) .. controls (146.28,102.37) and (143.02,105.43) .. (139.01,105.43) .. controls (134.99,105.43) and (131.74,102.37) .. (131.74,98.6) -- cycle ;
\draw  [fill={rgb, 255:red, 208; green, 2; blue, 27 }  ,fill opacity=0.4 ] (102.87,128.58) .. controls (106.64,128.62) and (109.67,131.9) .. (109.63,135.91) .. controls (109.6,139.93) and (106.51,143.15) .. (102.74,143.12) .. controls (98.96,143.08) and (95.93,139.8) .. (95.97,135.79) .. controls (96.01,131.77) and (99.1,128.55) .. (102.87,128.58) -- cycle ;
\draw  [fill={rgb, 255:red, 74; green, 144; blue, 226 }  ,fill opacity=0.4 ] (157.39,77.7) .. controls (157.39,73.93) and (160.65,70.87) .. (164.66,70.87) .. controls (168.67,70.87) and (171.93,73.93) .. (171.93,77.7) .. controls (171.93,81.47) and (168.67,84.53) .. (164.66,84.53) .. controls (160.65,84.53) and (157.39,81.47) .. (157.39,77.7) -- cycle ;
\draw  [fill={rgb, 255:red, 245; green, 166; blue, 35 }  ,fill opacity=0.4 ] (71.88,82.52) .. controls (71.88,78.75) and (75.14,75.69) .. (79.15,75.69) .. controls (83.17,75.69) and (86.42,78.75) .. (86.42,82.52) .. controls (86.42,86.3) and (83.17,89.36) .. (79.15,89.36) .. controls (75.14,89.36) and (71.88,86.3) .. (71.88,82.52) -- cycle ;
\draw [fill={rgb, 255:red, 126; green, 211; blue, 33 }  ,fill opacity=0.4 ]   (108.65,86.94) -- (103.95,98.2) ;
\draw [fill={rgb, 255:red, 248; green, 231; blue, 28 }  ,fill opacity=0.4 ]   (104,111) -- (102.87,128.58) ;
\draw [fill={rgb, 255:red, 126; green, 211; blue, 33 }  ,fill opacity=0.4 ]   (128.32,87.75) -- (134.3,94.18) ;
\draw [fill={rgb, 255:red, 126; green, 211; blue, 33 }  ,fill opacity=0.4 ]   (100.96,66.85) -- (106.94,70.87) ;
\draw [fill={rgb, 255:red, 139; green, 87; blue, 42 }  ,fill opacity=0.4 ]   (117.2,37.92) -- (136.01,49.97) ;
\draw [fill={rgb, 255:red, 65; green, 117; blue, 5 }  ,fill opacity=0.4 ]   (144.57,94.18) -- (159.96,82.93) ;
\draw [fill={rgb, 255:red, 126; green, 211; blue, 33 }  ,fill opacity=0.4 ]   (83.86,78.1) -- (88.99,68.46) ;
\draw [fill={rgb, 255:red, 139; green, 87; blue, 42 }  ,fill opacity=0.4 ]   (93.69,55.6) -- (106.09,39.52) ;
\draw  [fill={rgb, 255:red, 139; green, 87; blue, 42 }  ,fill opacity=0.4 ] (99.36,203.91) .. controls (99.36,200.13) and (102.62,197.07) .. (106.63,197.07) .. controls (110.64,197.07) and (113.9,200.13) .. (113.9,203.91) .. controls (113.9,207.68) and (110.64,210.74) .. (106.63,210.74) .. controls (102.62,210.74) and (99.36,207.68) .. (99.36,203.91) -- cycle ;
\draw  [fill={rgb, 255:red, 126; green, 211; blue, 33 }  ,fill opacity=0.4 ] (82.26,232.84) .. controls (82.26,229.07) and (85.51,226.01) .. (89.53,226.01) .. controls (93.54,226.01) and (96.8,229.07) .. (96.8,232.84) .. controls (96.8,236.61) and (93.54,239.67) .. (89.53,239.67) .. controls (85.51,239.67) and (82.26,236.61) .. (82.26,232.84) -- cycle ;
\draw  [fill={rgb, 255:red, 126; green, 211; blue, 33 }  ,fill opacity=0.4 ] (101.07,248.92) .. controls (101.07,242.04) and (107.01,236.46) .. (114.33,236.46) .. controls (121.65,236.46) and (127.58,242.04) .. (127.58,248.92) .. controls (127.58,255.8) and (121.65,261.37) .. (114.33,261.37) .. controls (107.01,261.37) and (101.07,255.8) .. (101.07,248.92) -- cycle ;
\draw  [fill={rgb, 255:red, 208; green, 2; blue, 27 }  ,fill opacity=0.4 ] (92.52,275.44) .. controls (92.52,271.67) and (95.78,268.61) .. (99.79,268.61) .. controls (103.8,268.61) and (107.06,271.67) .. (107.06,275.44) .. controls (107.06,279.21) and (103.8,282.27) .. (99.79,282.27) .. controls (95.78,282.27) and (92.52,279.21) .. (92.52,275.44) -- cycle ;
\draw  [fill={rgb, 255:red, 74; green, 144; blue, 226 }  ,fill opacity=0.4 ] (127.58,269.01) .. controls (127.58,265.24) and (130.83,262.18) .. (134.85,262.18) .. controls (138.86,262.18) and (142.12,265.24) .. (142.12,269.01) .. controls (142.12,272.78) and (138.86,275.84) .. (134.85,275.84) .. controls (130.83,275.84) and (127.58,272.78) .. (127.58,269.01) -- cycle ;
\draw  [fill={rgb, 255:red, 74; green, 144; blue, 226 }  ,fill opacity=0.4 ] (153.23,248.11) .. controls (153.23,244.34) and (156.49,241.28) .. (160.5,241.28) .. controls (164.51,241.28) and (167.77,244.34) .. (167.77,248.11) .. controls (167.77,251.89) and (164.51,254.94) .. (160.5,254.94) .. controls (156.49,254.94) and (153.23,251.89) .. (153.23,248.11) -- cycle ;
\draw  [fill={rgb, 255:red, 245; green, 166; blue, 35 }  ,fill opacity=0.4 ] (67.72,252.93) .. controls (67.72,249.16) and (70.98,246.1) .. (74.99,246.1) .. controls (79.01,246.1) and (82.26,249.16) .. (82.26,252.93) .. controls (82.26,256.71) and (79.01,259.77) .. (74.99,259.77) .. controls (70.98,259.77) and (67.72,256.71) .. (67.72,252.93) -- cycle ;
\draw    (104.49,257.36) -- (99.79,268.61) ;
\draw    (94.23,270.21) -- (79.7,257.36) ;
\draw    (124.16,258.16) -- (130.14,264.59) ;
\draw    (96.8,237.26) -- (102.78,241.28) ;
\draw [fill={rgb, 255:red, 65; green, 117; blue, 5 }  ,fill opacity=0.4 ]   (140.41,264.59) -- (155.8,253.34) ;
\draw    (79.7,248.51) -- (84.83,238.87) ;
\draw [fill={rgb, 255:red, 139; green, 87; blue, 42 }  ,fill opacity=0.4 ]   (114.33,236.46) -- (106.63,210.74) ;
\draw  [fill={rgb, 255:red, 126; green, 211; blue, 33 }  ,fill opacity=0.4 ] (60.66,203.1) .. controls (60.66,199.33) and (63.91,196.27) .. (67.93,196.27) .. controls (71.94,196.27) and (75.2,199.33) .. (75.2,203.1) .. controls (75.2,206.88) and (71.94,209.93) .. (67.93,209.93) .. controls (63.91,209.93) and (60.66,206.88) .. (60.66,203.1) -- cycle ;
\draw    (89.53,226.01) -- (73,209) ;
\draw  [fill={rgb, 255:red, 245; green, 166; blue, 35 }  ,fill opacity=0.4 ] (84.78,301.26) .. controls (84.64,305.03) and (81.28,307.97) .. (77.27,307.82) .. controls (73.25,307.67) and (70.12,304.49) .. (70.25,300.72) .. controls (70.39,296.95) and (73.76,294.01) .. (77.77,294.16) .. controls (81.78,294.31) and (84.92,297.49) .. (84.78,301.26) -- cycle ;
\draw    (95.43,279.54) -- (82.44,295.14) ;
\draw    (191,80) -- (303,80) ;
\draw [shift={(305,80)}, rotate = 180] [color={rgb, 255:red, 0; green, 0; blue, 0 }  ][line width=0.75]    (10.93,-3.29) .. controls (6.95,-1.4) and (3.31,-0.3) .. (0,0) .. controls (3.31,0.3) and (6.95,1.4) .. (10.93,3.29)   ;
\draw    (188,253) -- (302,253.98) ;
\draw [shift={(304,254)}, rotate = 180.49] [color={rgb, 255:red, 0; green, 0; blue, 0 }  ][line width=0.75]    (10.93,-3.29) .. controls (6.95,-1.4) and (3.31,-0.3) .. (0,0) .. controls (3.31,0.3) and (6.95,1.4) .. (10.93,3.29)   ;
\draw    (234,279) .. controls (235.64,280.69) and (235.61,282.36) .. (233.92,284) .. controls (232.23,285.64) and (232.2,287.31) .. (233.83,289) .. controls (235.47,290.69) and (235.44,292.36) .. (233.75,294) .. controls (232.06,295.64) and (232.03,297.31) .. (233.66,299) .. controls (235.3,300.69) and (235.27,302.36) .. (233.58,304) .. controls (231.89,305.64) and (231.86,307.31) .. (233.49,309) -- (233.47,310.01) -- (233.47,310.01) .. controls (235.16,308.36) and (236.82,308.38) .. (238.47,310.07) .. controls (240.12,311.76) and (241.78,311.78) .. (243.47,310.13) .. controls (245.16,308.48) and (246.82,308.5) .. (248.47,310.19) .. controls (250.12,311.88) and (251.78,311.9) .. (253.47,310.26) .. controls (255.16,308.61) and (256.82,308.63) .. (258.47,310.32) .. controls (260.12,312.01) and (261.78,312.03) .. (263.47,310.38) .. controls (265.16,308.73) and (266.82,308.75) .. (268.47,310.44) .. controls (270.12,312.13) and (271.78,312.15) .. (273.47,310.5) .. controls (275.16,308.85) and (276.82,308.87) .. (278.47,310.56) .. controls (280.12,312.25) and (281.78,312.27) .. (283.47,310.63) .. controls (285.16,308.98) and (286.82,309) .. (288.47,310.69) .. controls (290.12,312.38) and (291.78,312.4) .. (293.47,310.75) .. controls (295.16,309.1) and (296.82,309.12) .. (298.47,310.81) .. controls (300.12,312.5) and (301.78,312.52) .. (303.47,310.87) .. controls (305.16,309.23) and (306.82,309.25) .. (308.47,310.94) .. controls (310.12,312.63) and (311.78,312.65) .. (313.47,311) .. controls (315.16,309.35) and (316.82,309.37) .. (318.47,311.06) .. controls (320.12,312.75) and (321.78,312.77) .. (323.47,311.12) .. controls (325.16,309.47) and (326.82,309.49) .. (328.47,311.18) .. controls (330.12,312.87) and (331.78,312.89) .. (333.47,311.25) .. controls (335.16,309.6) and (336.82,309.62) .. (338.47,311.31) .. controls (340.12,313) and (341.78,313.02) .. (343.47,311.37) .. controls (345.16,309.72) and (346.82,309.74) .. (348.47,311.43) .. controls (350.12,313.12) and (351.78,313.14) .. (353.47,311.49) .. controls (355.16,309.85) and (356.82,309.87) .. (358.46,311.56) .. controls (360.11,313.25) and (361.77,313.27) .. (363.46,311.62) .. controls (365.15,309.97) and (366.81,309.99) .. (368.46,311.68) .. controls (370.11,313.37) and (371.77,313.39) .. (373.46,311.74) .. controls (375.15,310.09) and (376.81,310.11) .. (378.46,311.8) .. controls (380.11,313.49) and (381.77,313.51) .. (383.46,311.86) .. controls (385.15,310.22) and (386.81,310.24) .. (388.46,311.93) .. controls (390.11,313.62) and (391.77,313.64) .. (393.46,311.99) .. controls (395.15,310.34) and (396.81,310.36) .. (398.46,312.05) .. controls (400.11,313.74) and (401.77,313.76) .. (403.46,312.11) .. controls (405.15,310.46) and (406.81,310.48) .. (408.46,312.17) .. controls (410.11,313.86) and (411.77,313.88) .. (413.46,312.24) .. controls (415.15,310.59) and (416.81,310.61) .. (418.46,312.3) .. controls (420.11,313.99) and (421.77,314.01) .. (423.46,312.36) .. controls (425.15,310.71) and (426.81,310.73) .. (428.46,312.42) .. controls (430.11,314.11) and (431.77,314.13) .. (433.46,312.48) .. controls (435.15,310.84) and (436.81,310.86) .. (438.46,312.55) .. controls (440.11,314.24) and (441.77,314.26) .. (443.46,312.61) .. controls (445.15,310.96) and (446.81,310.98) .. (448.46,312.67) .. controls (450.11,314.36) and (451.77,314.38) .. (453.46,312.73) .. controls (455.15,311.08) and (456.81,311.1) .. (458.46,312.79) .. controls (460.11,314.48) and (461.77,314.5) .. (463.46,312.85) .. controls (465.15,311.21) and (466.81,311.23) .. (468.46,312.92) .. controls (470.11,314.61) and (471.77,314.63) .. (473.46,312.98) .. controls (475.15,311.33) and (476.81,311.35) .. (478.46,313.04) .. controls (480.11,314.73) and (481.77,314.75) .. (483.46,313.1) .. controls (485.15,311.45) and (486.81,311.47) .. (488.45,313.16) .. controls (490.1,314.85) and (491.76,314.87) .. (493.45,313.23) .. controls (495.14,311.58) and (496.8,311.6) .. (498.45,313.29) .. controls (500.1,314.98) and (501.76,315) .. (503.45,313.35) .. controls (505.14,311.7) and (506.8,311.72) .. (508.45,313.41) .. controls (510.1,315.1) and (511.76,315.12) .. (513.45,313.47) .. controls (515.14,311.83) and (516.8,311.85) .. (518.45,313.54) .. controls (520.1,315.23) and (521.76,315.25) .. (523.45,313.6) .. controls (525.14,311.95) and (526.8,311.97) .. (528.45,313.66) .. controls (530.1,315.35) and (531.76,315.37) .. (533.45,313.72) .. controls (535.14,312.07) and (536.8,312.09) .. (538.45,313.78) .. controls (540.1,315.47) and (541.76,315.49) .. (543.45,313.84) -- (546,313.88) -- (554,313.98) ;
\draw [shift={(556,314)}, rotate = 180.71] [color={rgb, 255:red, 0; green, 0; blue, 0 }  ][line width=0.75]    (10.93,-3.29) .. controls (6.95,-1.4) and (3.31,-0.3) .. (0,0) .. controls (3.31,0.3) and (6.95,1.4) .. (10.93,3.29)   ;
\draw    (241,62) .. controls (239.37,60.3) and (239.41,58.63) .. (241.11,57) .. controls (242.81,55.37) and (242.85,53.7) .. (241.22,52) .. controls (239.59,50.3) and (239.62,48.63) .. (241.32,47) .. controls (243.02,45.37) and (243.06,43.7) .. (241.43,42) -- (241.47,40.01) -- (241.47,40.01) .. controls (243.15,38.36) and (244.82,38.37) .. (246.47,40.05) .. controls (248.12,41.73) and (249.79,41.75) .. (251.47,40.1) .. controls (253.15,38.45) and (254.82,38.47) .. (256.47,40.15) .. controls (258.12,41.83) and (259.79,41.84) .. (261.47,40.19) .. controls (263.15,38.54) and (264.82,38.56) .. (266.47,40.24) .. controls (268.12,41.92) and (269.79,41.94) .. (271.47,40.29) .. controls (273.15,38.64) and (274.82,38.65) .. (276.47,40.33) .. controls (278.12,42.01) and (279.79,42.03) .. (281.47,40.38) .. controls (283.15,38.73) and (284.82,38.75) .. (286.47,40.43) .. controls (288.12,42.11) and (289.79,42.12) .. (291.47,40.47) .. controls (293.15,38.82) and (294.82,38.84) .. (296.47,40.52) .. controls (298.12,42.2) and (299.79,42.22) .. (301.47,40.57) .. controls (303.15,38.92) and (304.82,38.93) .. (306.47,40.61) .. controls (308.12,42.29) and (309.79,42.31) .. (311.47,40.66) .. controls (313.15,39.01) and (314.82,39.03) .. (316.47,40.71) .. controls (318.12,42.39) and (319.79,42.4) .. (321.47,40.75) .. controls (323.15,39.1) and (324.82,39.12) .. (326.47,40.8) .. controls (328.12,42.48) and (329.79,42.5) .. (331.47,40.85) .. controls (333.15,39.2) and (334.82,39.21) .. (336.47,40.89) .. controls (338.12,42.57) and (339.79,42.59) .. (341.47,40.94) .. controls (343.15,39.29) and (344.82,39.31) .. (346.47,40.99) .. controls (348.12,42.67) and (349.79,42.68) .. (351.47,41.03) .. controls (353.15,39.38) and (354.82,39.4) .. (356.47,41.08) .. controls (358.12,42.76) and (359.79,42.77) .. (361.47,41.12) .. controls (363.15,39.47) and (364.82,39.49) .. (366.47,41.17) .. controls (368.12,42.85) and (369.79,42.87) .. (371.47,41.22) .. controls (373.15,39.57) and (374.82,39.58) .. (376.47,41.26) .. controls (378.12,42.94) and (379.79,42.96) .. (381.47,41.31) .. controls (383.15,39.66) and (384.82,39.68) .. (386.47,41.36) .. controls (388.12,43.04) and (389.79,43.05) .. (391.47,41.4) .. controls (393.15,39.75) and (394.82,39.77) .. (396.47,41.45) .. controls (398.12,43.13) and (399.79,43.15) .. (401.47,41.5) .. controls (403.15,39.85) and (404.82,39.86) .. (406.47,41.54) .. controls (408.12,43.22) and (409.79,43.24) .. (411.47,41.59) .. controls (413.15,39.94) and (414.82,39.96) .. (416.47,41.64) .. controls (418.12,43.32) and (419.79,43.33) .. (421.47,41.68) .. controls (423.15,40.03) and (424.82,40.05) .. (426.47,41.73) .. controls (428.12,43.41) and (429.79,43.43) .. (431.47,41.78) .. controls (433.15,40.13) and (434.82,40.14) .. (436.47,41.82) .. controls (438.12,43.5) and (439.79,43.52) .. (441.47,41.87) .. controls (443.15,40.22) and (444.82,40.24) .. (446.47,41.92) .. controls (448.12,43.6) and (449.79,43.61) .. (451.47,41.96) .. controls (453.15,40.31) and (454.82,40.33) .. (456.47,42.01) .. controls (458.12,43.69) and (459.78,43.71) .. (461.46,42.06) .. controls (463.14,40.41) and (464.81,40.42) .. (466.46,42.1) .. controls (468.11,43.78) and (469.78,43.8) .. (471.46,42.15) .. controls (473.14,40.5) and (474.81,40.51) .. (476.46,42.19) .. controls (478.11,43.87) and (479.78,43.89) .. (481.46,42.24) .. controls (483.14,40.59) and (484.81,40.61) .. (486.46,42.29) .. controls (488.11,43.97) and (489.78,43.98) .. (491.46,42.33) .. controls (493.14,40.68) and (494.81,40.7) .. (496.46,42.38) .. controls (498.11,44.06) and (499.78,44.08) .. (501.46,42.43) .. controls (503.14,40.78) and (504.81,40.79) .. (506.46,42.47) .. controls (508.11,44.15) and (509.78,44.17) .. (511.46,42.52) .. controls (513.14,40.87) and (514.81,40.89) .. (516.46,42.57) .. controls (518.11,44.25) and (519.78,44.26) .. (521.46,42.61) .. controls (523.14,40.96) and (524.81,40.98) .. (526.46,42.66) .. controls (528.11,44.34) and (529.78,44.36) .. (531.46,42.71) .. controls (533.14,41.06) and (534.81,41.07) .. (536.46,42.75) .. controls (538.11,44.43) and (539.78,44.45) .. (541.46,42.8) .. controls (543.14,41.15) and (544.81,41.17) .. (546.46,42.85) .. controls (548.11,44.53) and (549.78,44.54) .. (551.46,42.89) -- (553,42.91) -- (561,42.98) ;
\draw [shift={(563,43)}, rotate = 180.53] [color={rgb, 255:red, 0; green, 0; blue, 0 }  ][line width=0.75]    (10.93,-3.29) .. controls (6.95,-1.4) and (3.31,-0.3) .. (0,0) .. controls (3.31,0.3) and (6.95,1.4) .. (10.93,3.29)   ;
\draw   (278.99,128.7) .. controls (282.75,128.74) and (285.76,131.81) .. (285.73,135.57) -- (285.27,182.97) .. controls (285.24,186.72) and (282.16,189.74) .. (278.41,189.7) -- (258.01,189.51) .. controls (254.25,189.47) and (251.24,186.4) .. (251.27,182.64) -- (251.73,135.24) .. controls (251.76,131.49) and (254.84,128.47) .. (258.59,128.51) -- cycle ;
\draw  [dash pattern={on 0.84pt off 2.51pt}]  (312,94) .. controls (331.2,81.52) and (283.12,114.2) .. (270.4,123.91) ;
\draw [shift={(269,125)}, rotate = 321.34000000000003] [color={rgb, 255:red, 0; green, 0; blue, 0 }  ][line width=0.75]    (10.93,-3.29) .. controls (6.95,-1.4) and (3.31,-0.3) .. (0,0) .. controls (3.31,0.3) and (6.95,1.4) .. (10.93,3.29)   ;
\draw  [dash pattern={on 0.84pt off 2.51pt}]  (307,243) .. controls (299.44,237.33) and (276.7,212.01) .. (270.84,200.81) ;
\draw [shift={(270,199)}, rotate = 428.2] [color={rgb, 255:red, 0; green, 0; blue, 0 }  ][line width=0.75]    (10.93,-3.29) .. controls (6.95,-1.4) and (3.31,-0.3) .. (0,0) .. controls (3.31,0.3) and (6.95,1.4) .. (10.93,3.29)   ;
\draw  [dash pattern={on 0.84pt off 2.51pt}]  (314,165) -- (469,165.99) ;
\draw [shift={(471,166)}, rotate = 180.36] [color={rgb, 255:red, 0; green, 0; blue, 0 }  ][line width=0.75]    (10.93,-3.29) .. controls (6.95,-1.4) and (3.31,-0.3) .. (0,0) .. controls (3.31,0.3) and (6.95,1.4) .. (10.93,3.29)   ;
\draw    (348.69,89.67) -- (473.69,154.67)(347.31,92.33) -- (472.31,157.33) ;
\draw    (348.78,246.57) -- (473.59,181.2)(350.18,249.23) -- (474.98,183.85) ;
\draw    (503.8,148.1) -- (567.8,62.1)(506.2,149.9) -- (570.2,63.9) ;
\draw    (500.32,177.28) -- (563.32,293.28)(497.68,178.72) -- (560.68,294.72) ;
\draw [fill={rgb, 255:red, 139; green, 87; blue, 42 }  ,fill opacity=0.4 ]   (127.58,248.92) -- (153.23,248.11) ;
\draw    (113,209) -- (155,243) ;
\draw  [fill={rgb, 255:red, 84; green, 150; blue, 13 }  ,fill opacity=0.4 ] (312,76.98) .. controls (312,68.16) and (320.95,61) .. (332,61) .. controls (343.05,61) and (352,68.16) .. (352,76.98) .. controls (352,85.81) and (343.05,92.96) .. (332,92.96) .. controls (320.95,92.96) and (312,85.81) .. (312,76.98) -- cycle ;
\draw  [fill={rgb, 255:red, 71; green, 119; blue, 175 }  ,fill opacity=0.4 ] (309.48,247.9) .. controls (309.48,239.07) and (318.43,231.92) .. (329.48,231.92) .. controls (340.53,231.92) and (349.48,239.07) .. (349.48,247.9) .. controls (349.48,256.73) and (340.53,263.88) .. (329.48,263.88) .. controls (318.43,263.88) and (309.48,256.73) .. (309.48,247.9) -- cycle ;
\draw  [fill={rgb, 255:red, 155; green, 75; blue, 7 }  ,fill opacity=0.4 ] (563,43) .. controls (563,34.17) and (571.95,27.02) .. (583,27.02) .. controls (594.05,27.02) and (603,34.17) .. (603,43) .. controls (603,51.83) and (594.05,58.98) .. (583,58.98) .. controls (571.95,58.98) and (563,51.83) .. (563,43) -- cycle ;
\draw  [fill={rgb, 255:red, 207; green, 145; blue, 40 }  ,fill opacity=0.4 ] (556,314) .. controls (556,305.17) and (564.95,298.02) .. (576,298.02) .. controls (587.05,298.02) and (596,305.17) .. (596,314) .. controls (596,322.83) and (587.05,329.98) .. (576,329.98) .. controls (564.95,329.98) and (556,322.83) .. (556,314) -- cycle ;
\draw  [fill={rgb, 255:red, 218; green, 233; blue, 134 }  ,fill opacity=0.4 ] (471,166) .. controls (471,157.17) and (479.95,150.02) .. (491,150.02) .. controls (502.05,150.02) and (511,157.17) .. (511,166) .. controls (511,174.83) and (502.05,181.98) .. (491,181.98) .. controls (479.95,181.98) and (471,174.83) .. (471,166) -- cycle ;

\draw (109,240.4) node [anchor=north west][inner sep=0.75pt]  [font=\large]    {$ \begin{array}{l}
v\\
\end{array}$};
\draw (113,70.4) node [anchor=north west][inner sep=0.75pt]   [font=\large]   {$ \begin{array}{l}
v\\
\end{array}$};
\draw (34,64.4) node [anchor=north west][inner sep=0.75pt]   [font=\large]   {$P_{1}$};
\draw (37,241.4) node [anchor=north west][inner sep=0.75pt]  [font=\large]    {$P_{2}$};
\draw (214,53.4) node [anchor=north west][inner sep=0.75pt]  [font=\large]    {$f_{1}$};
\draw (211,226.4) node [anchor=north west][inner sep=0.75pt]  [font=\large]    {$f_{2}$};
\draw (193,83) node [anchor=north west][inner sep=0.75pt] [font=\large]    [align=left] {encoder};
\draw (190,256) node [anchor=north west][inner sep=0.75pt]  [font=\large]   [align=left] {encoder};
\draw (320,64.4) node [anchor=north west][inner sep=0.75pt]  [font=\large]    {$\mathbf{z}^{P_{1}}_{v}$};
\draw (318,237.4) node [anchor=north west][inner sep=0.75pt]  [font=\large]    {$\mathbf{z}^{P_{2}}_{v}$};
\draw (574,31.4) node [anchor=north west][inner sep=0.75pt]   [font=\large]   {$\mathbf{s}^{P_{1}}$};
\draw (278.71,137.7) node [anchor=north west][inner sep=0.75pt]  [font=\large]  [rotate=-90.55] [align=left] {fusion};
\draw (294.01,155.91) node [anchor=north west][inner sep=0.75pt]   [font=\large]   {$g$};
\draw (485,160.4) node [anchor=north west][inner sep=0.75pt]  [font=\large]  {$\mathbf{z}_{v}$};
\draw (264,20) node [anchor=north west][inner sep=0.75pt]  [font=\large]   [align=left] {$c$ summarization};
\draw (264,318) node [anchor=north west][inner sep=0.75pt]  [font=\large]   [align=left] {$c$ summarization};
\draw (566,302.4) node [anchor=north west][inner sep=0.75pt]   [font=\large]   {$\mathbf{s}^{P_{2}}$};
\draw (382.39,82.55) node [anchor=north west][inner sep=0.75pt]  [font=\large]  [rotate=-28.15] [align=left] {consensus};
\draw (380.59,115.08) node [anchor=north west][inner sep=0.75pt]  [font=\large]  [rotate=-27.87]  {$I_{\psi _{1}}$};
\draw (362.21,214.86) node [anchor=north west][inner sep=0.75pt]  [font=\large]  [rotate=-333.03] [align=left] {consensus};
\draw (400.87,228.94) node [anchor=north west][inner sep=0.75pt]  [font=\large]  [rotate=-332.75]  {$I_{\psi _{2}}$};
\draw (493.98,122.83) node [anchor=north west][inner sep=0.75pt]  [font=\large]  [rotate=-309.43] [align=left] {consensus};
\draw (539.56,115.23) node [anchor=north west][inner sep=0.75pt]  [font=\large]  [rotate=-323.44]  {$I_{\phi _{1}}$};
\draw (533.92,192.19) node [anchor=north west][inner sep=0.75pt]  [font=\large]  [rotate=-60.09] [align=left] {consensus};
\draw (512.59,232.08) node [anchor=north west][inner sep=0.75pt]  [font=\large]  [rotate=-27.87]  {$I_{\phi _{2}}$};
\draw (396.09,67.42) node [anchor=north west][inner sep=0.75pt][font=\large] [rotate=-28.75] [align=left] {fine-grain};
\draw (358.44,195.46) node [anchor=north west][inner sep=0.75pt] [font=\large] [rotate=-335.12] [align=left] {fine-grain};
\draw (477.29,114.11) node [anchor=north west][inner sep=0.75pt] [font=\large] [rotate=-309.9] [align=left] {coarse-grain};
\draw (546.18,176.91) node [anchor=north west][inner sep=0.75pt] [font=\large] [rotate=-59.99] [align=left] {coarse-grain};

\end{tikzpicture}

%% file: figs/lambda.tex
\begin{tikzpicture}
\begin{axis}[legend style={at={(.9,0.9),anchor=north east}},
             legend style={legend pos=outer north east,}, xtick={0,.25,.5,.75,1}, xticklabels={0,.25,.5,.75,1}, title={\huge ACM}, ymin=93.28-1.6,ymax=93.28+1.6, yticklabel style = {font=\Large},
            xticklabel style = {font=\Large}
]

\addplot[mark=diamond*,violet,error bars/.cd,
            y dir=both, y explicit,] coordinates {
    (0,93.28)
    (.25,93.25)
    (.5,93.25)
    (.75,93.87)
    (1, 93.97)
    
};
\addplot[mark=diamond*,blue] coordinates {
    (0,93.29)
    (.25,93.39)
    (.5,93.26)
    (.75,93.90)
    (1, 94.07)
    
};

\end{axis}
\end{tikzpicture}

\begin{tikzpicture}
\begin{axis}[legend style={at={(.9,0.9),anchor=north east}},
             legend style={cells={anchor=east}, legend pos=outer north east,}, xtick={0,.25,.5,.75,1}, xticklabels={0,.25,.5,.75,1}, title={\huge DBLP},
             ymin=92.27-1.6,ymax=92.27+1.6,yticklabel style = {font=\Large},
            xticklabel style = {font=\Large},
]
\addplot[mark=diamond*,thick,violet] coordinates {
    (0,92.94)
    (.25,92.72)
    (.5,93.06)
    (.75,92.27)
    (1, 91.70)
    
};
\addplot[mark=diamond*,thick,blue] coordinates {
    (0,92.40)
    (.25,92.26)
    (.5,92.65)
    (.75,91.70)
    (1, 91.09)
    
};

\end{axis}
\end{tikzpicture}

\begin{tikzpicture}
\begin{axis}[legend style={at={(.9,0.9),anchor=north east}},
             legend style={cells={anchor=west}, legend pos=outer north east,}, xtick={0,.25,.5,.75,1}, xticklabels={0,.25,.5,.75,1}, title={\huge IMDB},
             ymin=62.97-1.6,ymax=62.9+1.6, yticklabel style = {font=\Large},
            xticklabel style = {font=\Large}
]

\addplot[mark=diamond*,thick,violet] coordinates {
    (0,63.49)
    (.25,63.65)
    (.5,63.41)
    (.75,63.23)
    (1, 62.97)
};
\addlegendentry{\large Micro-F1}
\addplot[mark=diamond*,thick,blue] coordinates {
    (0,63.48)
    (.25,63.75)
    (.5,63.22)
    (.75,63.35)
    (1, 62.90)
};
\addlegendentry{\large Macro-F1}
\end{axis}
\end{tikzpicture}

%% file: figs/atts.tex
\begin{tikzpicture}

\begin{axis} [ybar,xmin=0,xmax=1,ymin=0,ymax=1, enlarge x limits = {abs = .1},xtick = data, title={\huge ACM}]
\addplot coordinates {(0,0.6153) (0.25,0.6349) (0.5, 0.6812) (0.75, 0.7753) (1, 0.6299)};
\addplot coordinates {(0,0.3847) (0.25,0.3651) (0.5, 0.3188) (0.75, 0.2247) (1, 0.3701)};
\legend {PAP, PSP};
\end{axis}
\end{tikzpicture}

\begin{tikzpicture}
\begin{axis} [ybar,xmin=0,xmax=1,ymin=0,ymax=1, enlarge x limits = {abs = .1},xtick = data, title={\huge DBLP}]
\addplot coordinates {(0,0.0591) (0.25,0.0910) (0.5, 0.0783) (0.75, 0.0001) (1, 0.5196)};
\addplot coordinates {(0,0.1219) (0.25,0.0847) (0.5, 0.0481) (0.75, 0.3986) (1, 0.0002)};
\addplot coordinates {(0,0.8190) (0.25,0.8243) (0.5, 0.8736) (0.75, 0.6059) (1, 0.4810)};
\legend {APA, APTPA, APCPA};
\end{axis}
\end{tikzpicture}

\begin{tikzpicture}
\begin{axis} [ybar,xmin=0,xmax=1,ymin=0,ymax=1, enlarge x limits = {abs = .1},xtick = data, title={\huge IMDB}]
\addplot coordinates {(0,0.1245) (0.25,0.1081) (0.5, 0.1084) (0.75, 0.1684) (1, 0.5171)};
\addplot coordinates {(0,0.4096) (0.25,0.4137) (0.5, 0.4194) (0.75, 0.3989) (1, 0.2625)};
\addplot coordinates {(0,0.4659) (0.25,0.4782) (0.5, 0.4722) (0.75, 0.4327) (1, 0.2205)};
\legend {MDM, MAM, MKM};
\end{axis}

\end{tikzpicture}